\documentclass[journal]{IEEEtran}
\usepackage{hhline}
\usepackage{multirow, makecell}
\usepackage{array, booktabs}
\usepackage{bm, amsmath}
\usepackage{caption}
\usepackage[font={small}]{caption}
\usepackage[linesnumbered,ruled,vlined]{algorithm2e}
\usepackage{algpseudocode}
\usepackage{graphicx}
\usepackage{balance}
\usepackage{amssymb}
\usepackage{diagbox}
\usepackage{dblfloatfix}
\usepackage{tikz}
\usepackage{xcolor}

\newcommand{\YH}[1]{{\textcolor{black}{#1}}}

\makeatletter
\def\endthebibliography{%
	\def\@noitemerr{\@latex@warning{Empty `thebibliography' environment}}%
	\endlist
}
\makeatother

\begin{document}


\title{Scenario-Transferable Semantic Graph Reasoning for Interaction-Aware Probabilistic Prediction}

\author{Yeping~Hu, ~\IEEEmembership{Member,~IEEE,}
        Wei~Zhan, ~\IEEEmembership{Member,~IEEE,} \\
        and~Masayoshi~Tomizuka,~\IEEEmembership{Life Fellow,~IEEE}
\thanks{* Work done during Yeping's Ph.D. at University of California, Berkeley.}
\thanks{Y. Hu, W, Zhan, and M. Tomizuka are with the Department of Mechanical Engineering, University of California, Berkeley, Berkeley, CA 94720 USA (e-mail: yeping\_hu@berkeley.edu; wzhan@berkeley.edu; tomizuka@berkeley.edu).}}

\maketitle 
\begin{abstract}
Accurately predicting the possible behaviors of traffic participants is an essential capability for autonomous vehicles. Since autonomous vehicles need to navigate in dynamically changing environments, they are expected to make accurate predictions regardless of where they are and what driving circumstances they encountered. Several methodologies have been proposed to solve prediction problems under different traffic situations. These works usually combine agent trajectories with either color-coded or vectorized high definition (HD) map as input representations and encode this information for behavior prediction tasks. However, not all the information is relevant in the scene for the forecasting and such irrelevant information  may  be  even distracting to the forecasting in certain situations. Therefore, in this paper, we propose a novel generic representation for various driving environments by taking the advantage of semantics and domain knowledge. Using semantics enables situations to be modeled in a uniform way and applying domain knowledge filters out unrelated elements to target vehicle's future behaviors. We then propose a general semantic behavior prediction framework to effectively utilize these representations by formulating them into spatial-temporal semantic graphs and reasoning internal relations among these graphs. We theoretically and empirically validate the proposed framework under highly interactive and complex scenarios, demonstrating that our method not only achieves state-of-the-art performance, but also processes desirable zero-shot transferability.


\end{abstract}

\IEEEpeerreviewmaketitle

\section{Introduction}

\IEEEPARstart{P}{rediction} plays important roles in many fields such as economics \cite{economic}, weather forecast \cite{weather_forecast}, and human-robot interactions \cite{TRO_prediction_interaction}. For intelligent robots such as autonomous vehicles, accurate behavioral prediction of their surrounding entities could help them evaluate their situations in advance and drive safely. 
\begin{figure}[htbp]
	\centering
	\includegraphics[scale=0.54]{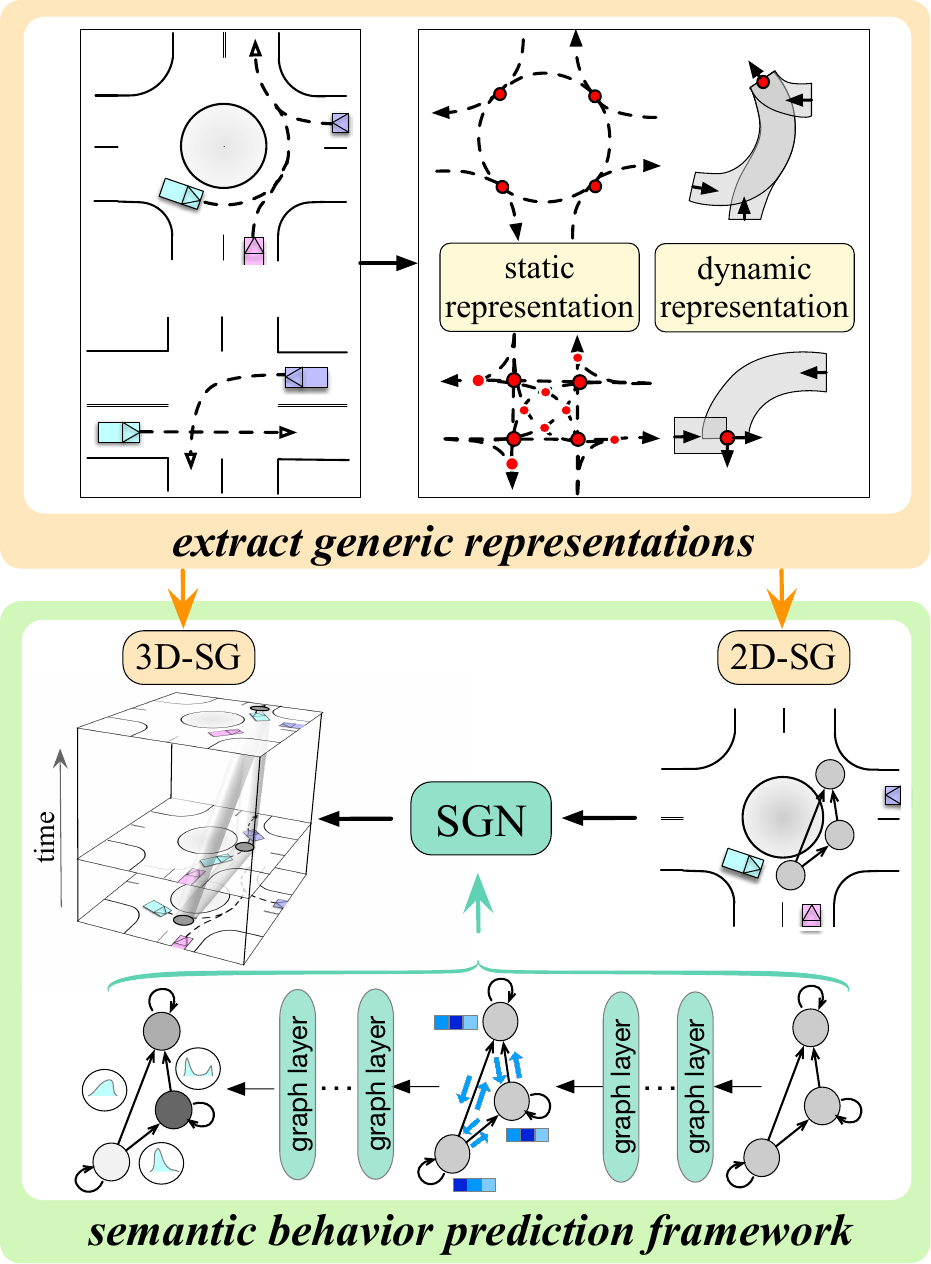}
	\caption{This paper introduces a generic representation extraction module and a semantic behavior prediction framework. Given any driving scenario, we first extract its generic static and dynamic representations using semantics and domain knowledge. We then introduce semantic graphs (SG) to support spatial-temporal structural relations within generic representations related to semantic goals. Finally, we utilize semantic graph network (SGN) to operate on semantic graphs and make predictions by reasoning internal structural relations of these graphs.}
	\label{fig:first_image}
\end{figure}



One challenge of developing prediction algorithms is to \textbf{\textit{find comprehensive and generic representations}} for common scenarios that can be encountered in the real world. Generic  representations  of  driving  environments  can  be regarded as invariant features across different driving scenarios or domains. In fact, finding suitable representations of the environment has been an open problem not only for prediction but also for decision making \cite{hu2019interaction} and planning \cite{TITS_planning_review} tasks. According to \cite{generalizable_intersect, transfer_intersect_ped_2}, and \cite{wang2020incorporating}, extracting invariant representations enhances generalizability for prediction models, which is extremely important for enabling autonomous vehicles to navigate in dynamically changing environment in real life. Many behavior prediction works built representations through color-coded attributes of the high definition (HD) map and trajectory information \cite{chauffeurnet, topology_traj_LSTM} . Other works such as \cite{waymo} and \cite{walters2020trajectory} proposed to directly utilize vectorized scene context and agent dynamics as representations. However, not all the information is relevant in the scene for the forecasting and such irrelevant information may be even distracting to the forecasting in certain situations. In this paper, we propose a novel semantic-based generic representation for driving environments while integrating domain knowledge. An advantage of using semantic approach is situations can be modeled in a unified way \cite{semantic} such that varying driving scenarios will have no effect on our semantics defined problem. Moreover, by considering domain knowledge, scene elements that are unrelated to or independent of future behaviors of target vehicles can be eliminated.


After generic representations are defined, another challenging problem arises as how to \textbf{\textit{effectively utilize these representations to predict human behaviors}}. 
Although deep learning methods have shown promising capabilities on highly data-driven problems \cite{TITS_prediction_data, TITS_big_data}, in order to achieve the best performance, designing or selecting a suitable network structure for given representations is still of great importance. For example, \cite{chauffeurnet} used Convolutional Neural Networks (CNNs) to encode rendered road maps, which can learn inherent properties of images and have better performance than classical neural networks. In contrast, \cite{waymo} applied Graph Neural Networks (GNNs) to incorporate the set of vectorized HD maps and agent trajectories, which can avoid lossy rendering and computationally intensive CNN encoding steps. In this paper, we propose a semantic behavior prediction framework for semantic representations. Because of the strong relational inductive biases and invariance of input ordering, the graph network structure is incorporated into our model. However, different from \cite{waymo} and other graph-based prediction approaches \cite{NRI, stochasticGNN, GNN_multi_agent_2}, the input and output of our model are semantic-based representations instead of agent trajectories or scene maps. Therefore, we specifically design a semantic graph reasoning process to infer internal spatial-temporal structural relations for our proposed semantic-based representations. By comparing with other state-of-the-art learning-based methods, we demonstrate that our model structure yields the best performance in terms of prediction accuracy. 

To the best of our knowledge, we are the first to propose semantic-based generic representations for driving environments based on domain knowledge, and we consequently propose a semantic behavior prediction framework to effectively utilize these representations (see Figure~\ref{fig:first_image}). The key contributions of this work are as follows:


\begin{itemize}
	\item We introduce generic representations for both static and dynamic driving environments by taking the advantage of semantics and domain knowledge.
	\item We propose a general semantic behavior prediction framework, which utilizes generic representations to define semantic goals and formulates input/output as 2D/3D semantic graphs (SG) based on semantic goals.
	\item We instantiate the general framework and propose a novel graph-based model for semantic behavior prediction (named SGN). We theoretically and empirically validate the proposed framework under highly interactive and complex scenarios, demonstrating that our method not only achieves state-of-the-art performance, but also processes desirable zero-shot transferability.
\end{itemize}

\section{Preliminaries and Problem Statement}

\subsection{Terminologies and Definitions}

\subsubsection{Domain knowledge}
For vehicle prediction problems, domain knowledge usually comprises information relate to traffic rules and traffic regulations. In this work, we further consider domain knowledge in terms of road topological constraints which can be regarded as hard attention mechanisms \cite{hard_attention} but without learnable parameters. These constrains directly filter out on-road agents that have no potential influences on target vehicle's behavior, which force the model to only pay attention to the relevant information while discard the others entirely to reduce information redundancy. More details on how domain knowledge is utilized during generic representation extraction can be found in Section \ref{static_env} and \ref{dynamic_env}.



\subsubsection{Semantic goals}
\label{semantic_goals}
These are goals that are described by semantics \cite{semantic}. To get a sense of what we mean by semantic goals, consider how a human driver think before performing any operations. These thoughts are not geometric — they do not take the form “reach the location that is 8.5 meters ahead and 2.4 meters to the left". Instead, the thoughts are of a semantic nature —  “cut in front of the blue car” or “stop behind the stop line” (see Figure \ref{fig:semantic_goal}). In Section \ref{dynamic_env}, we will explicitly illustrate how semantic goals are defined under our problem settings and how these goals can be modeled in a unified way regardless of driving environments. 
\subsubsection{Semantic Graph (SG)}
\label{semantic_graph}
We regard a graph \cite{original_GNN} as a semantic graph (SG) if its input and output features are defined by semantics. Two types of semantic graphs are introduced in this paper: two-dimensional semantic graph (2D-SG) and three-dimensional semantic graph (3D-SG). To be more specific, 2D-SG supports spatial relations among on-road entities and 3D-SG supports spatial-temporal structural relations related to semantic goals. Mathematical definitions of 2D-SG and 3D-SG can be found in Section \ref{SG}. 

\subsubsection{Semantic Graph Network (SGN)}
\label{semantic_graph_net}
The proposed semantic graph network (SGN) takes the advantage of the inductive biases in the graph network structure and operates on semantic graphs (SG). Specifically, SGN's input and output are represented by 2D-SG and 3D-SG respectively, and it makes predictions by reasoning internal structural relations of these graphs. Detailed structure of SGN is illustrated in Section \ref{SGN}.

\subsubsection{Zero-shot transferability}
It refers to the degree to which an existing model can be generalized or
transferred to various unseen scenarios or domains without updating model parameters. Detailed discussions and related works can be found in Section \ref{related works transfer}.

\begin{figure}[htbp]
	\centering
	\includegraphics[scale=0.55]{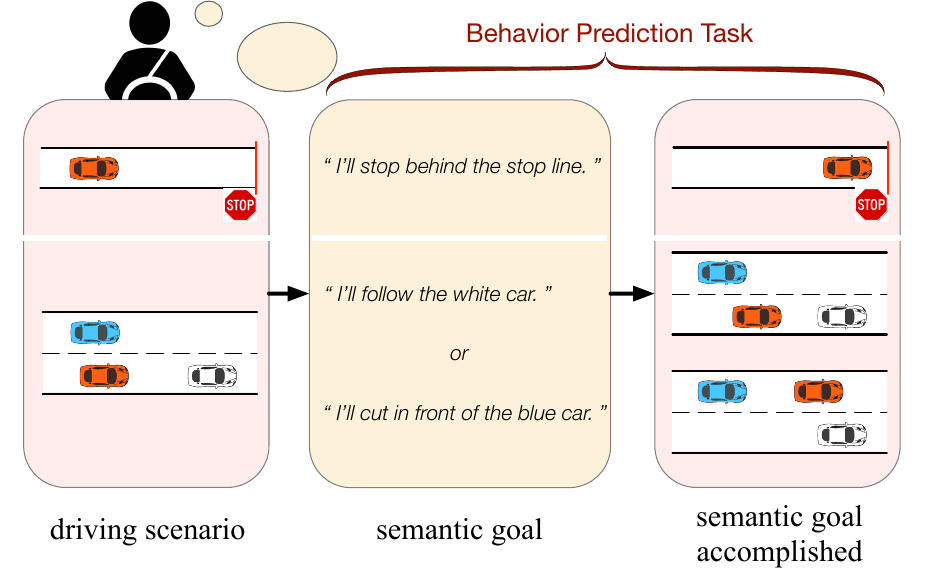}
	\caption{Illustration of the semantic goal concept as well as our behavior prediction task. The target vehicle is shown in red and our objective is to not only predict which semantic goal the driver is going to choose but also forecast corresponding vehicle's end state of achieving each semantic goal.}
	\label{fig:semantic_goal}
\end{figure}

\subsection{Problem Settings}
\label{problem_setting}
In this work, we focus on the goal-state prediction problem which contains both intention (e.g. left/right turn) and motion information (e.g. goal location and arrival time). In fact, by predicting goal states and assuming that agents navigate toward those goals by following some optimal or learned trajectories, the accuracy of prediction can be improved \cite{goal_plan_predestrain, precog}. For most goal-related prediction methods, the approach to selecting goals is either task-dependent \cite{Intentnet, ITSC_ttlc} (e.g. left/right turn) or by sampling-based method (e.g. along lane centerlines) \cite{TNT}. However, these potential goals are either too coarce to capture intra-category multimodality or based only on static map information without considering their dependencies with dynamically moving agents. Instead, we decide to use semantic goals in this work, which can cover all possible driving  situations as illustrated in Section \ref{semantic_goals}. 

In general, our objective is to predict the probability of each possible semantic goal and agent’s behaviors when each goal is accomplished (see Figure~\ref{fig:semantic_goal}). It is worth mentioning that the predicted outcomes can  be  further  used  for  goal-based  trajectory  estimation or planning tasks but it will not be the focus of this paper. Detailed problem statement can be found in Section \ref{problem_statement}.

\section{Related Works}
\label{related works}
In this section, we provide an overview of related works through three aspects and briefly discuss how each of them is addressed in this work. 

\subsection{Generic Representations of Driving Environments}
\label{related_work_representation}
Generic representations of driving environments can be regarded as invariant features across different driving scenarios or domains. Very few works have tried to find generic representations of driving environments. Jaipuria \textit{et al.} \cite{transfer_intersect_ped_2} applied affine transformation of pedestrians' trajectories into a uniform curbside coordinate frame. Hu \textit{et al.} \cite{hu_frenet_IRL}, \cite{hu2022causal} utilized the Fren\'{e}t coordinate frame along road reference paths to represent feature vectors of two interacting agents. In \cite{chauffeurnet}, self-centered image-based features were used as input to the network, where traffic regulations were encoded through images. Breuer \textit{et al.} \cite{topology_traj_LSTM} brought forward bird’s eye representation of the scene surrounding the object, fusing
various types of information on the scene which include satellite images
and bounding boxes of other traffic participants. 
A recent work from Waymo \cite{waymo} proposed a vectorized representation to encode HD maps and agent dynamics. Although such vectorized representations are applicable to various driving scenarios, even some simple or obvious relations between road or agent vectors have to be learned by the network and there is no guarantee that those known relations can be learned correctly. In fact, representations obtained from end-to-end deep learning models are with high abstraction level, which cannot be fully trusted and may fail under scenarios that are not well covered by the training data. 


\subsection{Behavior Prediction for Autonomous Vehicles}
\label{BP}
Many researchers have been focusing on probabilistic behavior prediction of autonomous vehicles. Methods such as deep neural networks (DNN) \cite{intersect_RNN_traj_ICRA, SIMP}, long short-term memory (LSTM) \cite{interact_LSTM_traj, social_LSTM}, convolutional neural networks (CNN) \cite{CNN_prediction, RAL_traj_LSTM_2}, conditional variational autoencoder (CVAE) \cite{CVAE_DESIRE}, generative adversarial network (GAN) \cite{social_GAN}, and gaussian process (GP) \cite{GP_traj} are typically utilized for trajectory prediction of intelligent agents.  Moreover, inverse optimal control (or inverse reinforcement learning (IRL)) method has also been use for probabilistic
reaction prediction under social interactions \cite{hu_frenet_IRL}, \cite{sun2019interpretable}. 

Recently, Graph Neural Network (GNN) has been widely used recently as it processes strong relational inductive biases \cite{original_GNN} and can achieve invariance to input ordering. In \cite{GNN_traj}, the authors utilized graph to represent the interaction among all close objects around the autonomous vehicle and employed an encoder-decoder LSTM model to make predictions. Instead of treating every surrounding agent equally, the attention mechanism was applied to GNN in \cite{GNN_att}, where the proposed graph attention network (GAT) can implicitly focus on the most relevant parts of the input (i.e. specify different weights to neighboring agents) to make decisions. Works such as \cite{NRI}, \cite{stochasticGNN}, and \cite{GNN_multi_agent_2} applied such a method to predict future states of multiple agents while considering their mutual relations. 

\subsection{Transferability of Prediction Algorithms}
\label{related works transfer}
Researchers have developed various machine learning algorithms to enhance the accuracy of behavior prediction tasks for autonomous vehicles under one or more driving scenarios such as highway (e.g. \cite{SIMP}, \cite{hu2018framework}, and \cite{TITS_highway_pred}), intersection (e.g. \cite{intersect_RNN_traj_ICRA}, \cite{GP_traj}, and \cite{TITS_intersection_pred}), and roundabout (e.g. \cite{hu2019multi} and \cite{roundabout}). Although these machine learning algorithms provide excellent prediction performance, a key assumption underlying the remarkable success is that the training and test data usually follow similar statistics. Otherwise, when test domains are unseen (e.g. different road structure from training domains) or Out-of-Distribution (OOD) \cite{lu2022generalizability}, \cite{OOD}, the resulting train-test domain shift will lead to significant degradation in prediction performance. Incorporating data from multiple training domains somehow alleviates this issue \cite{DG_2}, however, this may not always be applicable as it can be overwhelming to collect data from all the domains, especially for the autonomous driving industry.

Therefore, it is important for the predictor to have zero-shot transferability or domain generalizability, where it can be robust to domain-shift without requiring access to any data from testing scenarios during training. In this work, we will demonstrate the zero-shot transferability of our prediction algorithm when limited training domains are available.

\section{Proposed Solution}
In this paper, we proposed to solve the prediction problem (see Section~\ref{problem_setting}) by tackling the two challenges mentioned in the introduction section. This section is a brief overview of our approach.
\subsection{Find Comprehensive and Generic Representation} Generic representations of driving environments can be regarded as invariant features across different driving scenarios or domains. By finding such invariant/generic reprsentation, the zero-shot transferability of a prediction model can be achieved (see Section~\ref{related works transfer} and \ref{SG_theory}). In fact, as stated in Section~\ref{related_work_representation}, many researchers \cite{transfer_intersect_ped_2,hu_frenet_IRL,chauffeurnet,topology_traj_LSTM} have managed to to find such representations, but they either focus on extracting representations for a specific type of driving scenario or applying end-to-end learning approaches to implicitly learn generic representations across different scenarios. 

Instead, in this work, we take the advantage of human domain knowledge while constructing desired generic representations for various driving environments. Since any driving environment contains both static (e.g. map) and dynamic (e.g. on-road vehicles) information that have totally different properties, we propose separate pipelines to extract generic representations for \textit{static} (see Section~\ref{static_env}) and \textit{dynamic} (see Section~\ref{dynamic_env}) information. We comprehensively illustrate how we are able to represent any driving environment using the proposed representations. Note that such representations are model-agnostic and can be directly used as input features for any behavior prediction model (see Section~\ref{experiment}). 

\subsection{Effectively Predict Human Behaviors Using Generic Representation} 
Although our proposed representations can be utilized under any prediction model, in order to achieve the best performance, designing or selecting a suitable network structure for given representations is still of great importance. One difference between our semantic graphs and graphs mentioned in Section~\ref{BP} is the implication of node attributes. Typical graphs define each node as a single agent with its dynamic features. In contrast, we define each node as a semantic goal based on proposed generic representations (Section~\ref{static_env} and \ref{dynamic_env}), which implicitly contains features related to scene context and dynamics of varying number of agents. Therefore, in our setting, the relational reasoning process becomes more complicated as both inter- and intra- relationship exists in the semantic graph.

In this paper, we specifically design a semantic graph reasoning process to infer internal spatial-temporal structural relations for our proposed semantic-based representations and theoretical analyze the representational power of the corresponding model (see Section~\ref{SGN_theory}). The reasoning process is operated within the proposed semantic graph network (SGN) which takes the advantage of the inductive biases in the graph network structure (see Section~\ref{overall_framework}). Moreover, we use real-world data to demonstrate the effectiveness of our model in terms of prediction accuracy and zero-shot transferability compared with other state-of-the-art methods (see Section~\ref{experiment}). 

\begin{figure}[htbp]
	\centering
	\includegraphics[scale=0.55]{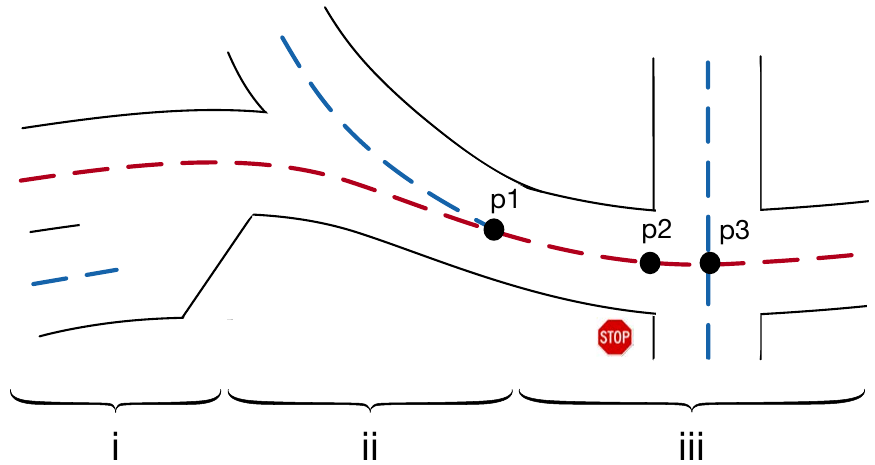}
	\caption{Illustration of reference paths (shown in dashed curves) and reference points (i.e. $p_1, p_2, p_3$) one of the paths (red). The lower-case roman numerals split the scenario into three different sections which represent various road topological relations.}
	\label{fig:ref_path}
\end{figure}
\section{Generic Representation of the Static Environment }
\label{static_env}
In order to design a prediction algorithm that can be used under different driving scenarios (e.g. highway, intersection, roundabout, etc.), we need a simple and generic representation of the static environment. The extracted expressions of the static environment should be able to describe road geometries and their interconnection as well as traffic regulations. We combine all these static environment information into road reference paths and the detailed methodology is described in this section.

\subsection{Reference Path}

A traffic-free reference path can be obtained either from road's centerline for constructed roads or by averaging human driving paths from collected data for unconstructed areas. The red and blue dashed lines in Fig.~\ref{fig:ref_path} denote different reference paths in the given scenario. 



\subsubsection{Reference point}
In order to incorporate map information into the reference path, we introduce the concept of \textit{reference points} which are selected points on the reference path. Reference points can either be \textbf{topological elements} that represent topological relations between two paths or \textbf{regulatory elements} that represent traffic regulations.

According to \cite{zhan2017spatially}, the topological relationship between any of two reference paths can be decomposed into three basic topological elements: \textit{point-overlap}, \textit{line-overlap}, and \textit{undecided-overlap}. In Fig.~\ref{fig:ref_path}, segment (i) has two parallel reference paths and can be categorized as undecided-overlap case corresponding to lane change or overtaking scenarios, which has no fixed reference point; segment (ii) is a merging scenario that belongs to the line-overlap case with reference point $p1$; segment (iii) is an intersection and can be regarded as point-overlap scenario with $p3$ as the reference point.

Moreover, a traveling path on public road is normally guided by regulatory elements like \textit{traffic lights} and \textit{traffic signs}. Therefore, it is reasonable to incorporate these regulatory elements into each reference path, where we utilize the reference point to denote the location of each regulatory element. As an example shown in Fig.~\ref{fig:ref_path}, the point $p2$ denotes the location of the stop line, which is one of the reference points on the red reference path.

\subsubsection{Mathematical definition}
We utilize the notation $\mathcal{X}_{ref}$ to represent the property of a reference path and each reference path is fitted by several way points through a polynomial curve and consists of various reference points. Therefore, we can mathematically define each reference path as $\mathcal{X}_{ref} = \{(x_k, y_k), (x_{p}, y_{p})\}$, where $x_{(\cdot)}$ and $y_{(\cdot)}$ are global locations of each point on the reference path, $k$ denotes the $k$-th way point, and $p$ denotes the $p$-th reference point. 

\subsection{Representation in Fren\'{e}t Frame}

In this work, we utilize the Fren\'{e}t Frame instead of Cartesian coordinate to represent the environment. The advantage of the Fren\'{e}t Frame is that it can utilize any selected reference path as the reference coordinate, where road geometrical information can be implicitly incorporated into the data without increasing feature dimensions. 
Specifically, given a vehicle moving on a reference path, we are able to convert its state from Cartesian coordinate ($x(t), y(t)$) into the longitudinal position $s(t)$ along the path, and lateral deviation $d(t)$ to the path. Note that the origin of the reference path is defined differently according to different objectives and each reference path will have its own Fren\'{e}t Frame. The following steps need to be performed throughout the testing process of the proposed architecture:
\begin{enumerate}
    \item At each time step, map the current global position (in Cartesian coordinate) of each vehicle on to each possible reference path, which results in position under Fren\'{e}t coordinate. 
    \item Perform the prediction algorithm under the
Fren\'{e}t coordinate.
    \item Convert the predicted results back to Cartesian coordinate to visualize the result.
\end{enumerate}

\section{Generic Representation of the Dynamic Environment }
\label{dynamic_env}
Based on the generic representation of the static environment defined in Section \ref{static_env}, we further design a uniform representation of the dynamic environment that can cover all types of driving situations on the road. In this section, we first redefine the Dynamic Insertion Area (DIA) concept, originally introduced in \cite{SIMP}, by providing comprehensive and mathematical definitions. We then thoroughly illustrate how the dynamic environment can be generically described by utilizing DIAs.

\subsection{Definition of DIA}
\subsubsection{General descriptions}
A dynamic insertion area (DIA) is semantically defined as: \textit{a dynamic area that can be inserted or entered by agents on the road}. An area is called dynamic when both its shape and location can change with time.  
\begin{figure}[htbp]
	\centering
	\captionsetup{justification=centering}
	\includegraphics[scale=0.37]{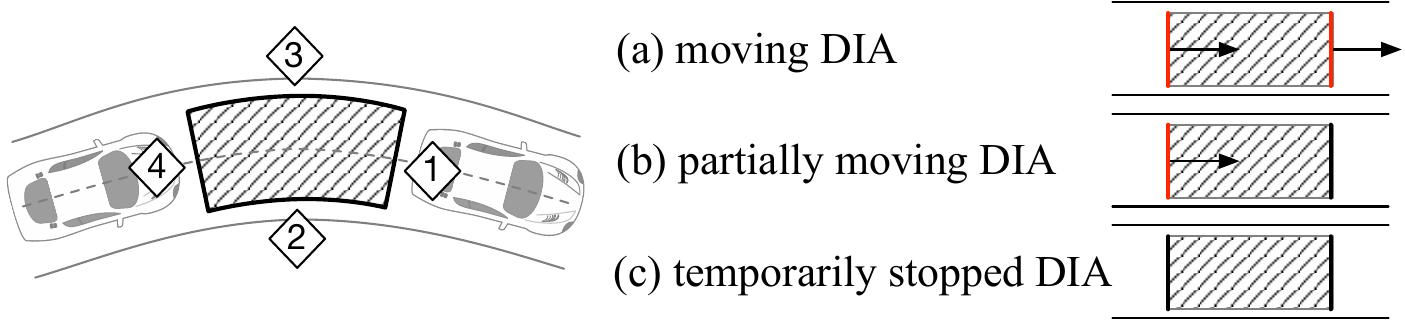}
	\caption{Basic properties for dynamic insertion area.}
	\label{fig:DIA}
\end{figure}
As can be seen in Fig.~\ref{fig:DIA}, each dynamic insertion area contains four boundaries: a front and a rear boundary (i.e. 1 \& 4), as well as two side boundaries (i.e. 2 \& 3). The front and rear boundaries of a DIA are usually formulated by road entities\footnote{The DIA boundaries can be formulated by any types of road entities including vehicles, cyclists and pedestrians. However, in this work, we will focus on vehicles only.}, but the two boundaries can also be any obstacles or predefined bounds based on traffic rules and road geometry. Since the two side boundaries for each DIA are formulated by connecting the front and rear boundary along road markings or curbs (as seen in Fig.~\ref{fig:DIA}), the shape of the area highly depends on the geometry of the reference path it is currently on. 



Each DIA has three different states as listed on the right side of Fig.~\ref{fig:DIA} and these states are categorized mainly by the motion of DIA's front and rear boundaries. For example, if both boundaries have non-zero speed, the corresponding DIA is called a moving DIA (i.e. Fig.~\ref{fig:DIA}(a)). If only one of the boundaries has zero speed, the DIA is regarded as partially moving (i.e. Fig.~\ref{fig:DIA}(b)). If, instead, both boundaries have zero speed, the DIA is temporarily stopped (i.e. Fig.~\ref{fig:DIA}(c)). Note that, in this work, we do not consider the case where the DIA is permanently stationary such as parking areas, which violates the dynamic property of DIA.

\subsubsection{Mathematical definition}
We define each dynamic insertion area as $\mathcal{A} = (\mathcal{X}_f, \mathcal{X}_r, \mathcal{X}_{ref})$, where $\mathcal{X}_f$ and $\mathcal{X}_r$ represent the properties of the front and rear bound of the DIA respectively; $\mathcal{X}_{ref}$, defined in the previous section, denotes the information of $\mathcal{A}$'s reference path which is the path that the area is currently moving on. Specifically, $\mathcal{X}_f = (x_f, y_f, v_f, a_f)$ and $\mathcal{X}_r = (x_r, y_r, v_r, a_r)$, where $x, y$ are the global locations of each boundary's center point, $v$ denotes the velocity, and $a$ denotes the acceleration. As the geometric properties of DIA's side boundaries can be described by $\mathcal{X}_{ref}$ and the states of each DIA mainly depend on its front and rear boundaries, we do not consider the states of side boundaries in the definition of $\mathcal{A}$. 

\begin{table}[ht]
	\caption{Features for the Dynamic Insertion Area}
	\label{tab:feature}
	\centering
	\begin{tabular}{c m{1.1cm} p{4.7 cm}}
		\toprule 
		& \textit{Feature} & \textit{Description} \\ 
		\midrule \midrule
		\multirow{2}{*}[-0.1cm]{\shortstack[lb]{\textbf{Area Spec}}}  & $l$ & Length of the area along reference path.\\ [1mm]
		& $\theta$ & Orientation of the area.\\
		\midrule
		\multirow{4}{*}[-0.8cm]{\shortstack[lb]{\textbf{Front/Rear} \\ \textbf{Boundary}}} & $v_{f/r}$ & Boundary's velocity in moving direction.\\ 
		& $a_{f/r}$ & Boundary's acceleration in moving direction. \\
		& $d^{lon}_{f/r}$ & Boundary's longitudinal distance to the active reference point. \\ 
		& $d^{lat}_{f/r}$ & Boundary's lateral deviation from the reference path. \\ 
		\bottomrule
	\end{tabular}
\end{table}

\subsubsection{Selected features}
As discussed in the previous section, we are able to utilize reference path $\mathcal{X}_{ref}$ as the reference coordinate under the Fren\'{e}t Frame and thus the property of each dynamic insertion area $\mathcal{A}$ can also be converted to the Fren\'{e}t Frame. We extract six higher-level features to represent each insertion area $\mathcal{A}$ from $(\mathcal{X}_f, \mathcal{X}_r, \mathcal{X}_{ref})$ under the Fren\'{e}t Frame, which are listed in Table~\ref{tab:feature}.

The length $l$ of each DIA is measured along its corresponding reference path, which can be expressed as: $l = d^{lon}_{f} - d^{lon}_{r}$. Here, $d^{lon}_{(\cdot)}$ denotes boundary's distance to the active reference point ${rpt}_{act}$ which is the point we select as the origin of the environment and might change with time. Note that for all DIAs in the scene that the predicted vehicle might reach, they will always share the same ${rpt}_{act}$ at a given time step. The criteria of choosing the active reference point will be discussed in the next subsection. The orientation $\theta$ of each area $\mathcal{A}$ is defined as the angle of the tangential vector to the reference path $\mathcal{X}_{ref}$ at the area's center point on the reference path, where the vector is pointed towards $\mathcal{A}$'s moving direction. Here, $\theta$ is measured relative to the global Cartesian coordinate instead of the local Fren\'{e}t Frame.

\subsection{DIAs in Dynamic Environment}
After introducing the basic concept of dynamic insertion area, we will first describe how to systematically extract DIAs in any given environments. We then illustrate how DIA can be combined with static environment information to generically represent different dynamic environments. Besides, we will examine the relationships amongst different DIAs when more than one DIA exists. As mentioned in Section \ref{static_env}, we are able to utilize two different aspects (i.e. topological and regulatory elements) to design a generic representation for the static environment. Therefore, we demonstrate the adaptability of DIA across several dynamical environments by varying the two aforementioned perspectives in the map (details shown in Fig.~\ref{fig:interact_DIA}). 

\subsubsection{Algorithm of extracting DIAs }

The entire algorithm of extracting DIAs in a scene, at any given time step, is described in Algorithm~\ref{alg:data_process_procedure}. The first step is to select the proper active reference point, the procedures of which are shown in Algorithm~\ref{alg:data_process_procedure}-(1). After obtaining the active reference point, we are able to extract all related DIAs in the environment by following the steps illustrated in Algorithm~\ref{alg:data_process_procedure}-(2). Since we are interested in the DIAs that the predicted vehicle might be inserted into, we only need to extract the DIAs within the observation range of the vehicle and we assume the predicted vehicle has full observation of its surroundings.

It is worth mentioning that it is possible for the predicted vehicle to have several possible reference paths when its high-level routing intention is ambiguous (e.g. the vehicle can either go straight or turn left/right at an intersection). In that case, Algorithm~\ref{alg:data_process_procedure} needs to be operated under each potential reference path of the predicted vehicle. However, since predicting high-level routing intention is not the focus of our interests in this work, without loss of generality, we assume that the predicted vehicle's ground-truth reference path is known throughout the rest of this paper. 

\begin{algorithm}
	\caption{Process of selecting the active reference point and extracting DIAs in a driving environment}
	\label{alg:data_process_procedure}
	
	\SetAlgoNoLine
	\SetArgSty{textup}

	For each predicted vehicle and the reference path $\mathcal{X}_{ref}$ it is moving on, do the following steps:\\

	(1) Find the active reference point ${rpt}_{act}$ in the scene: \\
	\quad$flag$ = False\Comment{${rpt}_{act}$ is not found yet}  \\

	\Indp\For{$\forall rpt$ (in front of the predicted vehicle) $\in$ $\mathcal{X}_{ref}$}{
		\If{$rpt \in $ \textit{regulatory elements}}{
			\If{$rpt \in $ \textit{traffic signs} \textbf{or} $rpt \in $ \textit{red traffic light}}{
			$flag$ = True \Comment{$rpt$ is $rpt_{act}$}\\
			break the loop
			}
		}
		\If{$rpt \in $ \textit{topological elements}}{
			\If{$\exists \mathcal{X}_{ref}'$ in the environment \textbf{s.t.} $\big(( \mathcal{X}_{ref}' \cap \mathcal{X}_{ref} = rpt )\wedge (\exists car \text{ on } \mathcal{X}_{ref}'))\big)$ }{
			$flag$ = True \Comment{$rpt$ is $rpt_{act}$}\\
			break the loop
			}
		}
	
	}
	\If(\Comment{no $rpt_{act}$ is found}){$flag$ == \text{False}}{ 
		define ${rpt}_{act}$ as a point in front of the predicted vehicle along $\mathcal{X}_{ref}$, with distance $d_{uo}$
	}

	\Indm (2) Find all DIAs in the scene up until ${rpt}_{act}$: \\
	\Indp\For{$\forall \mathcal{X}_{ref}'$ in the environment \textbf{s.t.} $\big(({rpt}_{act} \in \mathcal{X}_{ref}') \vee (\mathcal{X}_{ref}' \text{ is parallel to } \mathcal{X}_{ref})\big)$}{
		\eIf{$\mathcal{X}_{ref}' == \mathcal{X}_{ref}$}{
			 extract only the DIA in front of the predicted vehicle
			
		}{
			extract all DIAs along $\mathcal{X}_{ref}'$
		
		}

	}

\end{algorithm}
\subsubsection{DIAs under different topological relations}

We can use topological relations to represent the relationship between any two DIAs in a dynamic environment when they are moving on different reference paths, namely different $\mathcal{X}_{ref}$. The incorporations of DIAs under three basic topological elements are demonstrated as follows:\\

\begin{figure}[htbp]
	\centering
	\includegraphics[scale=0.5]{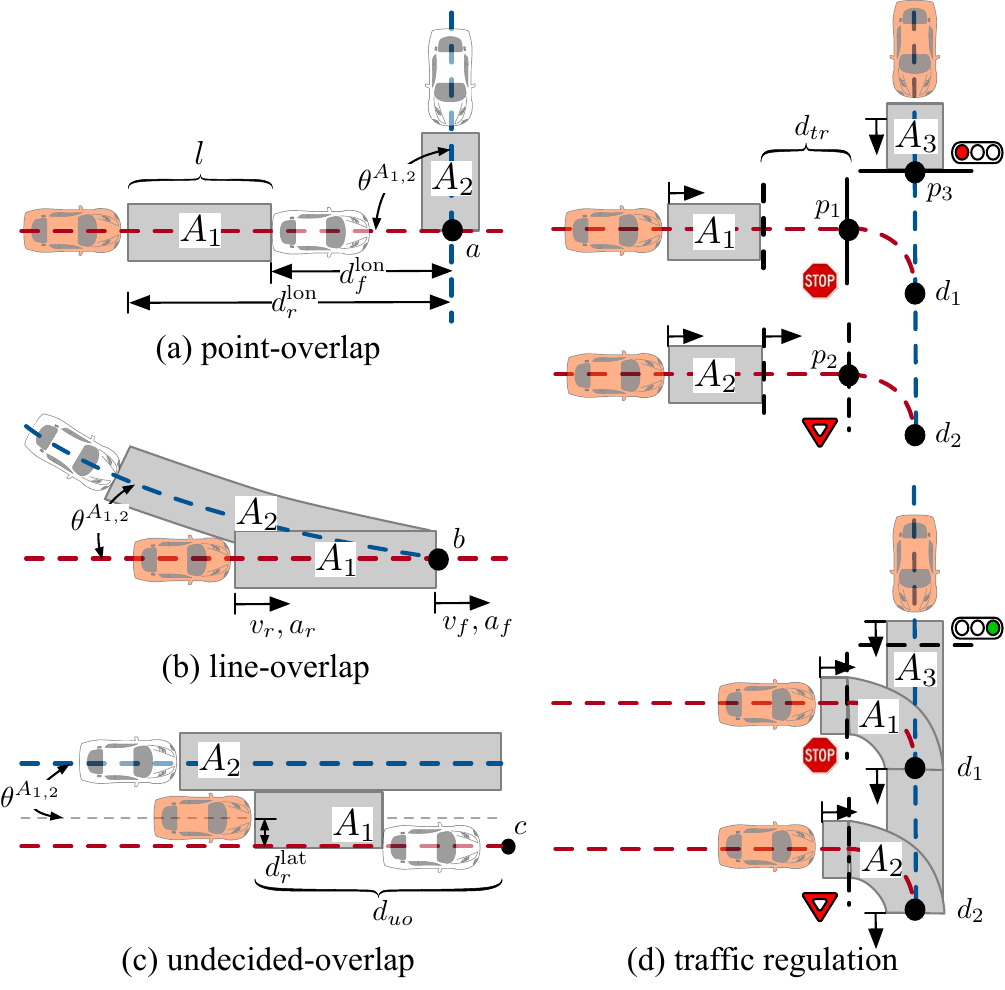}
	\caption{Demonstration of how DIA can be used to represent various driving environment. 
	}
	\label{fig:interact_DIA}
\end{figure}

\noindent\textbf{Point-overlap}: This corresponds to scenarios with crossing traffic such as intersections and an exemplar driving scenario is shown in Fig.~\ref{fig:interact_DIA}(a). When we consider the red car as the predicted vehicle, point $a$ is the active reference point in the scene according to the selection procedure in Algorithm~\ref{alg:data_process_procedure}. The two extracted DIAs are shaded in gray and their corresponding reference paths are represented in red (for $\mathcal{A}_1$) and blue (for $\mathcal{A}_2$) dashed lines. Hence, we denote the distances from the front bound and rear bound of $\mathcal{A}_1$ to $a$ as $d_f^{lon}$ and $d_r^{lon}$, respectively. The variable $\theta^{\mathcal{A}_{1,2}}$ denotes the relative angle between $\mathcal{A}_1$ and $\mathcal{A}_2$, where $\theta^{\mathcal{A}_{1,2}} = \theta^{\mathcal{A}_2} - \theta^{\mathcal{A}_1}$. In this example, both $\mathcal{A}_1$ and $\mathcal{A}_2$ are regarded as moving DIA. Here, we assign $\mathcal{A}_2$'s front boundary velocity $v^{\mathcal{A}_2}_f$ as the speed limit of its corresponding reference path (i.e. blue dashed line) and assume zero acceleration ($a^{\mathcal{A}_2}_f = 0$).\\

\noindent\textbf{Line-overlap}: This corresponds to scenarios of merging and car following, where an exemplar case is shown in Fig.~\ref{fig:interact_DIA}(b). In this situation, the front boundaries for $\mathcal{A}_1$ and $\mathcal{A}_2$ have some shared properties including $a_f$, $d^{lon}_f$, and $d^{lat}_f$. However, there front boundaries' velocities are not the same if their corresponding reference paths have different speed limits. \\

\noindent\textbf{Undecided-overlap}: This corresponds to scenarios that do not have a fixed merging or demerging point such as lane change. As can be seen in Fig.~\ref{fig:interact_DIA}(c), the two reference paths do not have a shared topological reference point that is fixed. For such situation, the active reference point in the scene is chosen to be the point, $c$, that has a pre-defined distance $d_{uo}$ to the predicted vehicle. Here, the dynamic insertion area $\mathcal{A}_1$ is moving towards $\mathcal{A}_2$ with a moving direction vertical to $\mathcal{A}_1$'s reference path. Therefore, the lateral deviation, $d^{lat}_r$, of $\mathcal{A}_1$'s rear bound is no longer closer to zero as in the previous two scenarios. Note that in order to clearly represent each DIA on the map, the front boundary of $\mathcal{A}_1$ shown in Fig.~\ref{fig:interact_DIA}(c) has the same lateral deviation as that of its rear boundary, however, the value of $d^{lat}_f$ should be consistent with the actual lateral deviation of $\mathcal{A}_1$'s  front boundary. Also, in this driving situation, the relative angle between two areas almost equals to zero (i.e. $\theta^{\mathcal{A}_{1,2}} \approx 0$). 

\subsubsection{DIAs under different traffic regulations}

As there can be several different traffic regulations that guide objects to move on each reference path, we categorize them into two groups and illustrate the incorporation of DIAs under each of these regulations:\\

	
\noindent\textbf{Traffic lights}: Traffic lights are usually positioned at road intersections, pedestrian crossings, and other locations to control traffic flows, which alternate the right of way accorded to road entities. If a reference path is guided by a traffic light, the reference point that represents such regulatory element is placed on the corresponding stop line. The signal color will affect the extraction of the dynamic insertion area. 
	For example, when we select the vehicle moving in the vertical direction as the predicted vehicle and the light in front of it is red (see the top scenario in Fig.~\ref{fig:interact_DIA}(d)), the active reference point is $p_3$. In such case, the stop line that $p_3$ is on is treated as the front boundary of $\mathcal{A}_3$, where $v^{\mathcal{A}_3}_f = 0$ and $\mathcal{A}_3$ is a partially moving DIA. When the light is green (see the bottom scenario in Fig.~\ref{fig:interact_DIA}(d)) or yellow, the active reference point for $\mathcal{A}_3$ switches to $d_1$. Under such situation, $v^{\mathcal{A}_3}_f$  equals to the speed limit on the blue reference path and thus $\mathcal{A}_3$ is regarded as a moving DIA. \\
	
\noindent\textbf{Traffic signs}: Traffic signs can be grouped into several types such as priority signs, prohibitory signs, and mandatory signs. In fact, sign groups that contain prohibitory and mandatory signs can be directly incorporated into the static environment representation by defining different reference paths. In this section, we only consider the sign groups that have influences on the dynamic environment. The sign group that is most commonly seen on the road is the groups of priority signs. Priority traffic signs include the stop and yield sign, which indicate the order in which vehicles should pass intersection points.
	
	\quad When a vehicle is moving towards a stop sign, it will first decrease its speed before reaching the stop line and then slowly inching forward while paying attention to other lanes. In order to represent the differences between these two stages through DIA, we create a virtual stop line at a distance $d_{tr}$ before the actual stop line. Fig.~\ref{fig:interact_DIA}(d) illustrates this two-stage process, where the active reference point for the vehicle behind the stop sign changes from $p_1$ to $d_1$ and the front boundary of $\mathcal{A}_1$ moves from the virtual stop line to the line across $d_1$ (see the transition from the top to the bottom scenario in Fig.~\ref{fig:interact_DIA}(d)). During the whole process,  $\mathcal{A}_1$ transforms from a partially moving DIA into a moving DIA. Alternatively, if a yield sign is encountered, the vehicle will not necessarily decrease its speed unless it has to yield other vehicles on the main path. However, if the speed limit on the yield path is lower than that of on the main path, the two-stage process is also necessary. Such situation is illustrated in Fig.~\ref{fig:interact_DIA}(d) where $\mathcal{A}_2$ remains as a moving DIA throughout the process. It is noteworthy that in Fig.~\ref{fig:interact_DIA}(d), when we separately predict the two horizontally moving vehicles, the front boundary for $\mathcal{A}_3$ will vary due to different selection of the active reference point (i.e. when the vehicle behind the stop sign is predicted, the front boundary of $\mathcal{A}_3$ is at $d_1$; when the vehicle behind the yield sign is predicted, $\mathcal{A}_3$'s front boundary changes to $d_2$).

\section{Semantic Behavior Prediction Framework}
\label{overall_framework}

In this section, we first state the prediction problem we aim to solve in this work. Then the concept of semantic graph (SG) is explained. Finally, we introduce the proposed semantic graph network (SGN) which can predict behaviors of interacting agents by reasoning their internal relations. The overall proposed semantic behavior prediction framework is shown in Figure \ref{fig:entire_framework}.


\subsection{Problem Statement}
\label{problem_statement}
As discussed in Section \ref{problem_setting}, we aim at directly predicting vehicle's behaviors related to its semantic goals. Specifically, we decide to use our proposed dynamic insertion area (DIA) as a uniform description of semantic goals and integrate it into the spatial-temporal semantic graph (SG) to construct a structural representation of the environment. In fact, as each DIA represents a semantic goal, we will use these two terminologies interchangeably in the rest of this paper. It is worth to address that the reason we regard DIA as semantics is twofold: (1) DIA is defined by semantic description; (2) navigation-relevant semantic map information can be explicitly or implicitly included in DIAs.

Therefore, in this work, we would like to predict or answer the following questions: \textbf{\textit{``Which DIA will the vehicle most likely insert 
into eventually? Where is the insertion location? When will the insertion take place?"}}.
\subsection{Semantic Graphs (SG)}
\label{SG}


The 2D-SG is defined similar to the traditional graph \cite{original_GNN} $\mathcal{G} = ( \mathcal{N}, \mathcal{E})$ with node $n \in \mathcal{N}$ and edge $e = (n, n') \in \mathcal{E}$ which represents a directed edge from $n$ to $n'$. For undirected edge , it can be modeled by explicitly assigning two directed edges in opposite directions between two nodes. The feature vector associated with node $n_i$ at time step $t$ is denoted as $\mathbf{x}_i^t$. The feature vector associated with edge $e_{ij} = (n_i, n_j)$ at time step $t$ is denoted as $\mathbf{x}_{ij}^t$. Note that within a 2D-SG, only spatial relations are described since different nodes are connected using edges at the same time-step.

Alternatively, we define a 3D-SG as $\mathcal{G}^{t\rightarrow t'} = (\mathcal{N}^{t\rightarrow t'}, \mathcal{E}^{t\rightarrow t'})$, where $t\rightarrow t'$ denotes the time span from time step $t$ to a future time step $t'$ with $t' > t$. The graph $\mathcal{G}^{t\rightarrow t'}$ contains information that spans the entire period of scene evolution. The spatial and temporal relationship are jointly described by edges in 3D-SG, where the temporal relation between any of the two nodes in a 3D-SG can differ. We define node $n^\tau \in \mathcal{N}^{t\rightarrow t'}$ with $\{\tau \in \mathbb{R} | t \leq \tau \leq t' \}$ and edge $e^{t\rightarrow t'} = (n^t, (n')^{t'}) \in \mathcal{E}^{t\rightarrow t'}$. The feature vector associated with node $n_i$ at time step $t_i$ is denoted as $\mathbf{x}_i^{t_i}$. The feature vector associated with edge $e_{ij}^{t_i \rightarrow t_j} = (n_i^{t_i}, n_j^{t_j})$ that spans from $t_i$ to $t_j$ is denoted as $\mathbf{x}_{ij}^{t_i \rightarrow t_j}$. Note that when $t_i = t_j$, the spatial-temporal edge is the same as the spatial edge in 2D-SG (i.e. $e_{ij}^{t_i \rightarrow t_j} = e_{ij}$ ).

\begin{figure}[htbp]
	\centering
	\includegraphics[scale=0.32]{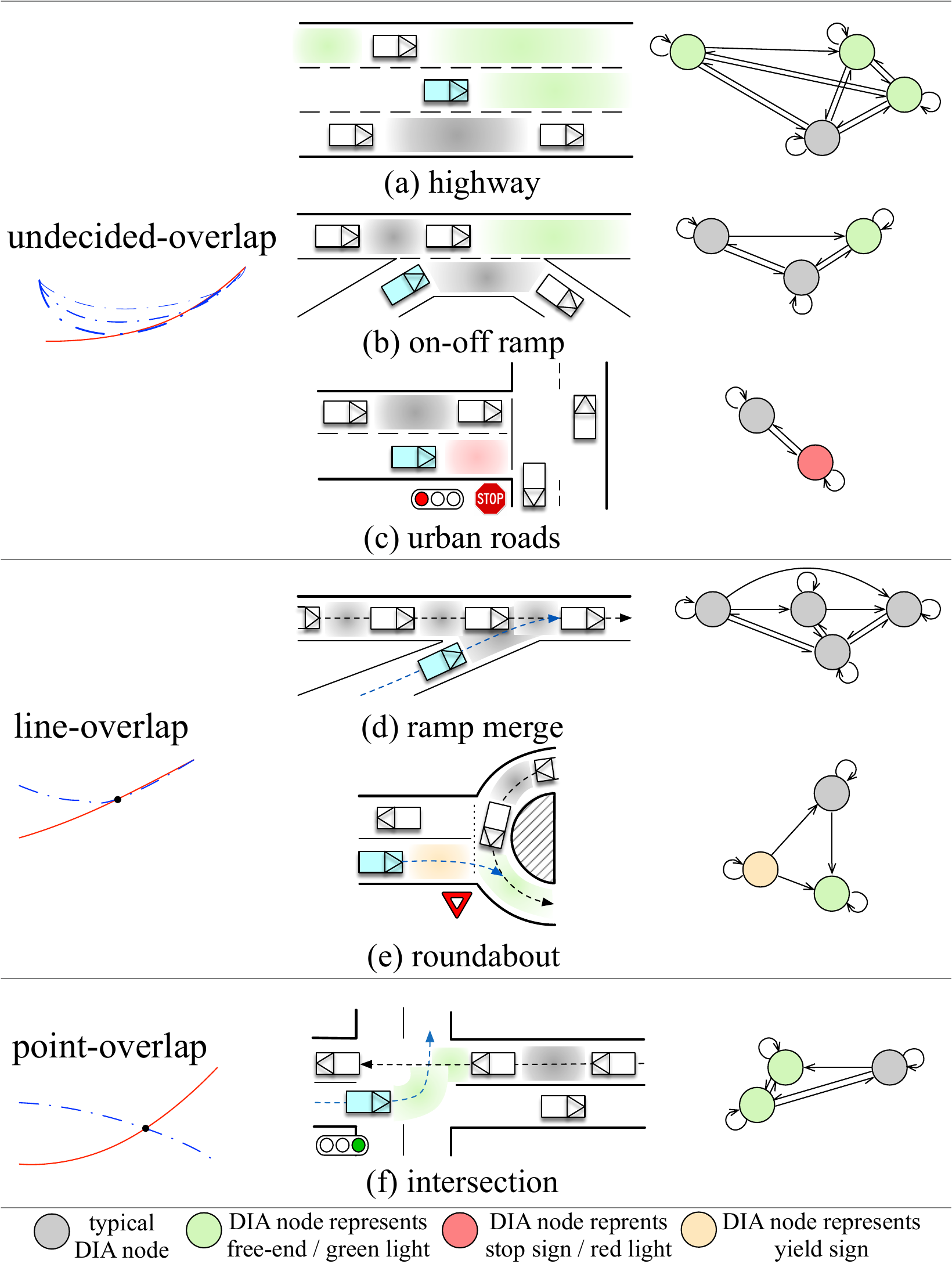}
	\caption{Illustration of various driving scenarios with extracted DIAs and corresponding 2D semantic graphs. The predicted vehicle is colored in cyan. Notice that all DIA nodes in the scene are defined uniformly and 
    we color the DIAs for better interpretation of different driving situations. }
	\label{fig:all_scene}
\end{figure}

For driving scenarios, rather than assigning node attribute as individual entities (e.g. vehicles), we utilize DIA instead. Since DIA can not only describe dynamic environments but also inherently incorporate static map information, each node in the graph is able to represent semantic information in the environment. By defining node attributes as semantic objects like DIA, we are able to implicitly encode both static and dynamic information into the graph. 
Moreover, the edge attribute describes the relationship between any of two DIAs. For a 2D-SG, each edge may describe the strength of correlation between its corresponding two DIAs at the same time step; whereas for a 3D-SG, each edge may represent some future information of the two DIAs. For example, in the 3D-SG of scene in Fig.~\ref{fig:interact_DIA}(c), the edge between $\mathcal{A}_1$ and $\mathcal{A}_2$ might encode the information of when and how two areas will merge together, which can be interpreted as when the red vehicle will cut in front of its left vehicle and at what location. 
Details on how various driving scenarios can be represented by semantic graphs are illustrated in Fig~\ref{fig:all_scene}.

\begin{figure*}[htbp]
	\centering
	\includegraphics[scale=0.5]{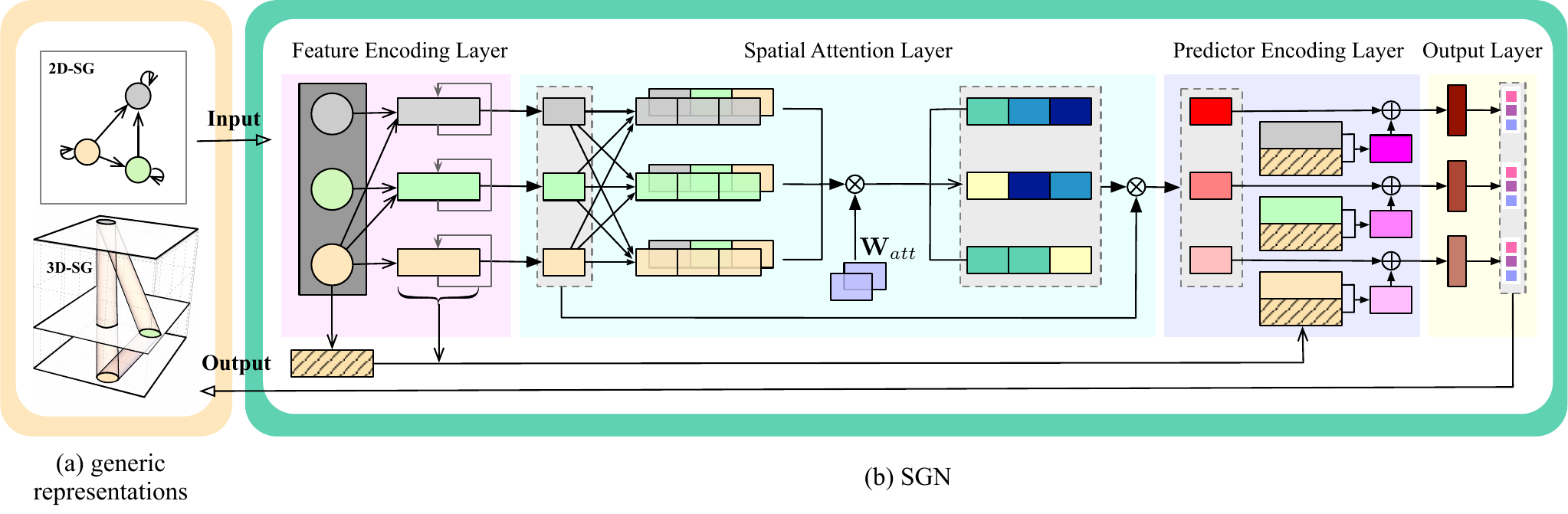}
	\caption{Illustration of the proposed semantic behavior prediction framework. (a) This general framework utilizes generic representations to define semantic goals and formulates input/output as 2D/3D semantic graphs (SG) based on semantic goals. (b) We further propose a semantic graph network (SGN) to reason internal spatial-temporal structural relations of semantic graphs and make predictions. SGN contains feature encoding layer, spatial attention layer, predictor encoding layer, and output layer.}
	\label{fig:entire_framework}
\end{figure*}

\subsection{Semantic Graph Network (SGN)}
\label{SGN}
The entire architecture of SGN is shown in Fig.~\ref{fig:entire_framework}(b) and each module is explained in details. 

\subsubsection{Input and output}
For the proposed framework, the input can either be a 2D-SG at the current time step or a sequence of historical 2D-SGs. The output is a set of 3D-SGs that encompass the information of how the current scene will progress in the future, which could provide answers to our questions in Section V.A.

\subsubsection{Feature encoding layer}
Since the state information of the predicted vehicle is implicitly contained in the DIA directly in front of it (i.e. the predicted vehicle forms the rear boundary of its front DIA),  we can regard its front DIA as the reference DIA in the scene. Therefore, in a 2D-SG, if node $i$ is selected as the reference DIA at the current time step $t$, we can then define the feature vector of some node $n_j$ relative to node $n_i$ as $\mathbf{x}_{j \rightarrow i}^t$, denoted as the relative node feature, and can be obtained by the following equation:
\begin{equation}
	\mathbf{x}_{j \rightarrow i}^t = f_{j \rightarrow i}(\mathbf{x}_i^t, \mathbf{x}_j^t),
\end{equation}
where $\mathbf{x}_i^t$ and $\mathbf{x}_j^t$ are absolute node features described in Section IV.A. We utilize a linear function $f_{j \rightarrow i}$ to map from absolute to relative node features. If we also have historical information of the node attribute, we can add a recurrent layer to further encode these sequential features:
\begin{equation}
	\mathbf{\hat{h}}_{j \rightarrow i}^t = f_{rec}^1(\mathbf{\hat{h}}_{j \rightarrow i}^{t-1}, \mathbf{x}_{j \rightarrow i}^t),
\end{equation}
where $\mathbf{\hat{h}}_{j \rightarrow i}^t$ is the hidden state also the output of the recurrent function $f_{rec}^1$. We choose the graph recurrent unit (GRU) \cite{GRU} as our recurrent function, where for each node $n_j$ the input to the GRU update is the previous hidden state $\mathbf{\hat{h}}_{j \rightarrow i}^{t-1}$ and the current input  $\mathbf{x}_{j \rightarrow i}^t$.

Similarly, we encode feature sequences of the reference DIA by applying another recurrent function $f_{rec}^2$:
\begin{equation}
	\mathbf{\hat{h}}_i^t = f_{rec}^2(\mathbf{\hat{h}}_i^{t-1}, \mathbf{x}_i).
\end{equation}

\subsubsection{Spatial attention layer}
The task of the attention layer is to help with modeling the locality of interactions among DIAs and improve performance by determining which DIAs will share information. We first encode the embedded features from the previous layer to yield a fixed-length vector $\mathbf{h}_{j \rightarrow i}^t$:
\begin{eqnarray}
	\mathbf{h}_{j \rightarrow i}^t = f_{enc}^1(\mathbf{\hat{h}}_{j \rightarrow i}^t),
\end{eqnarray}
where $f_{enc}^1$ denotes the encoder. We then compute attention coefficients $a_{(j \rightarrow i)(k \rightarrow i)}^t$ that indicate the importance of relative node feature $\mathbf{h}_{j \rightarrow i}^t$ to node feature $\mathbf{h}_{k \rightarrow i}^t$ as follows:
\begin{eqnarray}
	a_{jk}^t = f_{att}(concat(\mathbf{h}_{j \rightarrow i}^t, \mathbf{h}_{k \rightarrow i}^t); \mathbf{W}_{att}),
\end{eqnarray}
where we have simplified $a_{(j \rightarrow i)(k \rightarrow i)}^t$ as $a_{jk}^t$ for readability. We denote $\mathbf{W}_{att}$ as the attention weight and $f_{att}$ as a function that maps each concatenated two node intention features into a scalar. To make coefficients easily comparable across different node relations, we normalize them across all choices of $j$ using the \textit{softmax} function:
\begin{equation}
	\alpha_{jk}^t = \frac{\text{exp}(a_{jk}^t)}{\sum_{k' \in \mathcal{N}_i^t}\text{exp}(a_{jk'}^t)},
\end{equation} 
where $\alpha_{jk}^t$ denotes the normalized attention coefficient and $\mathcal{N}_i^t$ is a set of nodes that surrounds $n_i$ in the graph at time step $t$. Finally, we derive the attention-weighted relative node feature $\bar{\mathbf{h}}_{j \rightarrow i}^t$, which is an encoded vector weighted by attention as:
\begin{equation}\label{eq:aggregator_1}
	\bar{\mathbf{h}}_{j \rightarrow i}^t = \sum_{k' \in \mathcal{N}_i^t} \alpha_{jk'}^t \odot \mathbf{h}_{k' \rightarrow i}^t,
\end{equation}
where $\odot$ is the element-wise multiplication.

\subsubsection{Predictor Encoding layer}

For a given predicted vehicle, there will always be a DIA that is right in front of it and we regard this DIA as the reference DIA as stated previously. Therefore, if we want to infer the relations between the predicted vehicle and each of the DIAs on the road, we can alternatively infer the relations between the reference DIA and each of the other DIAs. Note that the predicted vehicle can also insert into the reference DIA (i.e. its front DIA), which might correspond to car-following in highway scenarios or yielding other cars in merging scenarios. 

Therefore, we need to encode the relationship between any of the two nodes and make a prediction on their relations in the future. Such predicted relations will be reflected through the edges in the output 3D-SG. For any pair of nodes $(i,j)$ that has connected edges in the input 2D-SG, we first concatenate their features to formulate the edge feature as either
\begin{eqnarray}
    \mathbf{\hat{h}}_{ij}^t = concat(\mathbf{\hat{h}}_{j \rightarrow i}^t, \mathbf{\hat{h}}_i^t) \quad \text{or} \quad 
     e_{ij}^t = concat(\mathbf{x}_{j \rightarrow i}^t, \mathbf{x}_i^t),
\end{eqnarray}
depending on whether we have embedded historical node features or not. $\mathbf{\hat{h}}_{ij}^t$ denotes the hidden edge relation between node $i$ and $j$ over certain past horizon. We can then generate an embedded vector $\mathbf{h}_{ij}^t$ as follows:
\begin{equation}
	\mathbf{h}_{ij}^t = f_{enc}^2(\mathbf{\hat{h}}_{ij}^t),
	\label{eq:f_enc_2}
\end{equation}
where the subscript $(\cdot)_{ij}$ denotes that $i$ is the index of the reference node and $j$ is the index of the node that connects to it. Different from the previous encoding function $f_{enc}^1$ that outputs encoded information for each node, the function $f_{enc}^2$ aims at encoding the edge information. After the edge encoding, we sum the result with the aggregated feature of the reference node, which aggregation scheme is inspired by \cite{GNN_theory1}, and perform a decoding process $f_{pred}$ to generate predicted edge information: 
\begin{equation}\label{eq:aggregator_2}
	\mathbf{o}_{ij} = f_{pred}((1 + \epsilon)\cdot\bar{\mathbf{h}}_{j \rightarrow i}^t + \mathbf{h}_{ij}^t),
\end{equation}
where $\epsilon$ can be either a learnable parameter or a fixed scalar, and $\mathbf{o}_{ij}$ denotes the encoded feature vector that will be later used to generate features for the 3D-SG.

\subsection{Output Layer}
In order to determine what elements should be generated by $\mathbf{o}_{ij}$, we first need to know what behavior we want to predict by revisiting our problem. Our task is to generate probabilistic distributions of the states for every possible insertion area in the input 2D-SG. In other word, we want to have a probabilistic distribution of states for every edge in the output 3D-SG. Without loss of generality, we assume the distribution can be described by a mixture of Gaussian. By using a mixture model, more flexibility can be given to model completely general conditional density function \cite{MDN}. In fact, bivariate Gaussian model has been widely used by researchers in many vehicle prediction models \cite{GNN_multi_agent_2, social_LSTM, graves2013generating} to approximate agent's next-state (i.e. position in 2D coordinate) distribution and has shown better performance than other density models. In this work, we extend it to multivariate Gaussian model as our output state is three-dimensional and we assign a Gaussian Mixture Model (GMM) to each 3D edge, where each Gaussian mixture models the probability distribution of a certain type of edge relation between its two connected nodes. 

To infer the final insertion location of the predicted vehicle, we need to have at least two predicted variables: location of the inserted DIA and location of the vehicle in that DIA. Besides, since the time of insertion is also the focus of our interests, a 3D Gaussian mixture is used and the predicted variables are constructed as a three dimensional vector: $\mathbf{y} = [y_{s_1}, y_{s_2}, y_t]^T$. The variable $y_{s_1}$ denotes the location of the inserted DIA, $y_{s_2}$ denotes the location of the predicted vehicle relative to the DIA it enters, and $y_t$ denotes the time left for the predicted vehicle to finish the insertion. 

Predicting when and where the predicted vehicle will be inserted into a particular DIA associated with node $j$, is equivalent to predict edge relations between the predicted vehicle's front DIA (assuming it is associated with node $i$) and $n_j$. Hence, given the encoded edge feature vector $\mathbf{o}_{ij}^{t}$, the probability distribution of the output $\mathbf{y}_{ij}^{t_i \rightarrow t_j}$ over the edge $e_{ij}^{t_i \rightarrow t_j}$ is of the form $f(\mathbf{y}_{ij}^{t_i \rightarrow t_j}| \mathbf{o}_{ij})$ . For brevity, we will eliminate the superscript of $\mathbf{y}_{ij}^{t_i \rightarrow t_j}$ for the rest of the paper.  Since we utilize the Gaussian kernel function to represent the probability density, we can rewrite $f(\mathbf{y}_{ij} |\mathbf{o}_{ij})$ as:
\begin{equation}
	\begin{split}
	f(\mathbf{y}_{ij} |\mathbf{o}_{ij})& = f\big(\mathbf{y}_{ij}|f_{out}^1(\mathbf{o}_{ij})\big)\\
	&= \sum_{m = 1}^{M}\alpha^{m}_{ij}\mathcal{N}\big(\mathbf{y}_{ij}|\bm{\mu}_{ij}^m, \Sigma_{ij}^{m}\big),
	\end{split}
	\label{eq:gaussian}
\end{equation}
where $\mathcal{N}\big(\mathbf{y}_{ij}|\bm{\mu}_{ij}^m, \Sigma_{ij}^{m}\big)$ can be expanded as:
\begin{equation}
\mathcal{N}\big(\mathbf{y}_{ij}|\bm{\mu}_{ij}^m, \Sigma_{ij}^{m}\big) = \frac{\text{exp}(-\frac{1}{2}(\mathbf{y}_{ij} - \mu_{ij}^m)^T\Sigma^{-1}(\mathbf{y}_{ij} - \mu_{ij}^m))}{\sqrt{(2\pi)^d |\Sigma|}},
\label{eq:pdf}
\end{equation}
where $d$ denotes the output dimension which is three in this problem. In Eq.~\ref{eq:gaussian}, $M$ denotes the total number of mixture components and the parameter $\alpha^m_{ij}$ denotes the $m$-th mixing coefficient of the corresponding kernel function. The function $f_{out}^1$ maps input $\mathbf{o}_{ij}$ to the parameters of the GMM (i.e. mixing coefficient $\alpha$, mean $\mu$, and covariance $\Sigma$), which in turn gives a full probability density function of the output $\mathbf{y}_{ij}$. Specifically, the mean and covariance are constructed as:
\begin{equation}
	\bm{\mu} = \begin{bmatrix}
	\mu_{s_1}\\\mu_{s_2}\\\mu_{t}
	\end{bmatrix},
	\Sigma = \begin{bmatrix}
	\sigma_{s_1}^2 & \sigma_{(s_1,s_2)} & \sigma_{(s_1, t)}\\
	\sigma_{(s_2, s_1)} & \sigma_{s_2}^2 & \sigma_{(s_2, t)} \\
	\sigma_{(t, s_1)} & \sigma_{(t, s_2)} & \sigma_{t}^2
	\end{bmatrix}.
\end{equation} 
Besides predicting the state of final insertion in each DIA for the predicted vehicle, we also want to know the probability of inserting into each DIA observed in the scene. Therefore, given the encoded edge feature vector $\mathbf{o}_{ij}^t$, we further derive the insertion probability of node $j$'s associated DIA as:
\begin{eqnarray}
	w_{ij} = \frac{1}{1 + exp(f_{out}^2(\mathbf{o}_{ij}^t))},
\end{eqnarray}
which is the \textit{logistic} function of the scalar output from function $f_{out}^2$. We also normalize the insertion probability such that $\sum_{k \in \mathcal{N}_i}w_{ik} = 1$.

Finally, we obtain the feature vector associated with each edge in a 3D-SG as: $\mathbf{x}_{ij}^{t_i \rightarrow t_j} = [\mathbf{y}_{ij}, w_{ij}]$. In the case where $i$ is the reference node, $t_i$ represents the current time of prediction and $t_j$ is sampled from the distribution of the predicted insertion time variable $y_t$. By sampling from the predicted distribution of each edge in 3D-SG, we are able to formulate several 3D-SGs as possible outcomes of the scene evolution.

\subsection{Loss Function}
For the desired outputs, we expect not only the largest
weight to be associated to the actual inserted area ($\mathcal{L}_{class}$), but also
the highest probability at the correct location and time for
the output distributions of that area ($\mathcal{L}_{regress}$). Consequently, our loss function becomes the summation of the negative log-likelihood loss of multivariate Gaussian distribution \cite{MDN, graves2013generating} and the cross-entropy loss:
\begin{equation} \label{eq:loss}
\begin{split}
\mathcal{L} &= \mathcal{L}_{regress} + \beta \mathcal{L}_{class}  \\ &= -\sum_{\mathcal{G}_s}\sum_{i \in \mathcal{N}^s}\bigg(\log\bigg\{\sum_{k \in \mathcal{N}_i^s}\hat{w}_{ik}	f(\mathbf{y}_{ik} |\mathbf{o}_{ik})\bigg\} \\
& \quad - \beta\sum_{k \in \mathcal{N}_i^s}\hat{w}_{ik}\log(w_{ik})\bigg),
\end{split}
\end{equation}
where $\mathcal{G}_s$ denotes the $s$-th 2D-SG, $\mathcal{N}^s$ denotes all the nodes in the current semantic graph, and $\mathcal{N}_i^s$ denotes the set of nodes surround $n_i$. Note that the number of nodes in set $\mathcal{N}^s$ and $\mathcal{N}_i^s$ is not fixed and will vary with time. The one-hot encoded ground-truth final inserting area is denoted by $\hat{w}_{ik}$. The hyperparameter $\beta$ is used to control the balance between the two losses for better performance.

\subsection{Design Details}
In this work, we utilize feed-forward neural networks for all related functions described in Secton \ref{SGN} (i.e. $f_{enc}^{1,2}$, $f_{att}$, $f_{pred}$, $f_{out}^{1,2}$), due to neural network's strong capacity of learning and modeling complex relationships between input and output variables. The function $f_{out}^1$ can thus be regarded as a GMM-based mixture density network (MDN) \cite{MDN}. It is important to note that the parameters of the GMM need to satisfy specific
conditions in order to be valid. For example, the mixing coefficients $\alpha^m$ should be positive and sum to 1 for all $M$, which can be satisfied by applying a \textit{softmax} function.
Also, the standard deviation for each output variable should be positive, which can be fulfilled by applying an exponential operator. 

Moreover, we should notice that in Eq.~\ref{eq:pdf}, $\Sigma$ is invertible only when it is a positive definite matrix. However, there is no guarantee that our formulated covariance matrix is non-singular. One solution to fix a singular covariance matrix is to create a new matrix $\hat{\Sigma} = \Sigma + kI$, where we want all the eigenvalues of the new covariance matrix be positive such that the matrix is invertible. Ideally, we prefer $k$ to be a very small number so not to bias our original covariance matrix. At the meantime, since the eigenvalues of $\hat{\Sigma}^{-1}$ are the reciprocals of the originals, we want $k$ to be large enough so that the eigenvalues of $\hat{\Sigma}^{-1}$ won't explode. Therefore, the best way of selecting $k$ is hyperparameter tuning.

\subsection{Inference for Semantic Prediction}
At test time, we fit the trained model to observed historical 2D-SGs up until the current time step $t$ and sample from the probabilistic density function $f(\mathbf{y}|\mathbf{o})$ to obtain a set of possible scene evolution outcomes. Although the network only output edge features of the 3D-SG, the node features are implicitly predicted as we know the spatial-temporal relations between any of two nodes. Hence, each sampled testing results can be formulated as a 3D-SG and we will thus obtain a set of 3D-SGs. 

It should be pointed out that for a given 2D-SG, if the reference DIA is changed, we might end up obtaining different output 3D-SGs. This is reasonable since as we modify the reference node, we potentially alter the vehicle we want to predict. Therefore, under the perspective of distinct drivers, the scene will evolve into the future differently. On the other hand, if we assume vehicle-to-vehicle (V2V) communication, it is possible for all drivers on the road to reach a consensus on the future states of the scene.
\YH{\section{Theoretical Analysis}}
\label{theory}
\YH{In this section, we theoretically evaluate the capability of the proposed algorithm.}

\subsection{Representation in Semantic Graph}
\label{SG_theory}
\YH{One of the main differences between our semantic graphs and typical graphs are the implication of node attributes. To be more specific, SG's node features are based on generic representations that are invariant to driving scenarios/domains, while typical graphs' node features are domain variant. Here, we show the advantage of using generic representation and semantic graphs.}

\YH{For real-world driving data, the joint distribution (i.e. $P(X, Y)$) between any two scenarios/domains can have discrepancies due to different map information or driving situations. In order to improve zero-shot transferability from source to target domain, our goal is to minimize the joint discrepancy across domains in a shared feature space and map the output policy on the shared features to minimize the target error. According to Bayes Theorem, $P(X, Y) = P(X)P(Y|X)$. Without loss of generality, by assuming the conditional distribution $P(Y|X)$ (i.e. the optimal predictor) is shared across domains, our goal becomes to minimize changes in marginal distribution $P(X)$.}

\noindent \YH{\textbf{Definition 1} (Marginal distribution alignment). \textit{Given two domains $D_S = \{(\mathbf{x}_i^S, \mathbf{y}_i^S)\}_{i=1}^{N_S}$ and $D_T = \{(\mathbf{x}_i^T, \mathbf{y}_i^T)\}_{i=1}^{N_T}$ drawn from two different distributions $\Pr(X^S, Y^S) \neq \Pr(X^T, Y^T)$ with non-zero covariant shift $\textrm{KL}(\Pr(X^S, Y^S)||\Pr(X^T, Y^T)) > 0$, marginal alignment corresponds to finding the feature transformation $g : \mathcal{X} \rightarrow \mathcal{Z}$ such that the discrepancy between the transformed marginal distribution is minimized, i.e., $\Pr(g(X^S)) = \Pr(g(X^T))$}.}

\YH{In fact, theoretical insights of zero-shot transferable algorithms are based on the hypothesis that a provable low target error can be obtained by minimizing the marginal discrepancies between two classifiers \cite{transfer_theory1, transfer_theory2}.}

\noindent \YH{\textbf{Theorem 1} (Ben-David et al. \cite{transfer_theory1})\textbf{.} \textit{Given two domains $D_s$ and $D_T$, the error of a hypothesis $\pi \in \mathcal{H}$ in the target domain $\epsilon_T(\pi)$ is bounded by the sum of: 1) the error of the hypothesis in the source domain, 2) the marginal discrepancy of the hypothesis class between the domains $d_{\mathcal{H}\Delta \mathcal{H}}(D_S^Z, D_T^Z):= 2\sup_{\pi,\pi'\in\mathcal{H}}|\epsilon_S(\pi, \pi') - \epsilon_T(\pi, \pi')|$, and 3) the best-in-class joint hypothesis error $\lambda_\mathcal{H} = \min_{\pi\in\mathcal{H}}[\epsilon_S(\pi)+\epsilon_H(\pi)]$},}
\begin{equation}
    \YH{\epsilon_T(\pi) \leq \epsilon_S(\pi) + \frac{1}{2}d_{\mathcal{H}\Delta \mathcal{H}}(D_S^Z, D_T^Z) + \lambda_\mathcal{H}.}
\end{equation}

\YH{Therefore, in this work, we propose to align marginal discrepancy across domains by finding generic representations of driving environment, where the feature transformation $g$ in Definition 1 equivalents to the generic representation extraction process described in Section~\ref{static_env} and \ref{dynamic_env}. By utilizing such representation as graph features, the proposed semantic graph has shown to have similar marginal distribution across domains (see Section IX-F). In this way, zero-shot transferability can be achieved for a given predictor using proposed generic representations as input features according to Theorem 1. }

\subsection{Design of Semantic Graph Network}
\label{SGN_theory}
\YH{In order to theoretically explain the idea behind our model design, we utilize a theoretical framework proposed in \cite{GNN_theory1} that can analyze the representation power of GNNs. The framework represents the set of feature vectors of a given node's neighbors as a \textit{multiset}, \textit{i.e.}, a set with possibly repeating elements. The formal definition is as follows:} 

\noindent\YH{\textbf{Definition 2} (Multiset). A multiset if a generalized concept of a set that allows multiple instances for its elements. More formally, a multiset is a 2-tuple $X = (S, m)$ where $S$ is the \textit{underlying set} of $X$ that is formed from its distinct elements, and $m : S \rightarrow \mathbb{N}_{\geq 1}$ gives the multiplicity of the elements.}

\YH{The neighbor aggregation in GNNs can be then thought of as an \textit{aggregation function over the multiset}. The more discriminative the multiset function is, the more powerful the representational power of the underlying GNN. According to \cite{GNN_theory1}, a maximally powerful GNN would \textit{never} map two nodes, \textit{i.e.,} multisets of feature vectors, to the same representation in the embedding space. This means its aggregation scheme must be \textit{injective}: }

\noindent\YH{\textbf{Theorem 2} (Xu et al. \cite{GNN_theory1}). \textit{Let $\mathcal{A} : \mathcal{G} \rightarrow \mathbb{R}^d$ be a GNN. With a sufficient number of GNN layers, $\mathcal{A}$ maps any graphs $G_1$ and $G_2$ that the Weisfeiler-Lehman test of isomorphism decides as non-isomorphic, to different embeddings if $\mathcal{A}$ aggregates and updates node features iteratively with}}
\YH{\begin{equation}
    h_v^{(k)} = \phi(h_v^{(k-1)}, f(\{h_u^{k-1}: u \in \mathcal{N}(v)\})),
\end{equation}}
\YH{\textit{where $ h_v^{(k)}$ is the feature vector of node $v$ at the $k$-th layer, the functions $f$, which operates on multisets, and $\phi$ are injective. }}
\YH{Moreover, the following theorem \cite{GNN_theory2} proves that the number of independent aggregators used in a GNN is a limiting factor of its expressive power and learning abilities.} \setlength{\parskip}{5pt}

\noindent\YH{\textbf{Theorem 3} (Number of aggregators needed)\textbf{.} \textit{In order to discriminate between multisets of size $n$ whose underlying set is $\mathbb{R}$, at least $n$ aggregators are needed.}}

\YH{Therefore, in order to design a GNN that has desirable representation power, we need to have multiple injective aggregators equivalent to the number of multiset. Most work in the literature uses only a single aggregation method with \textit{mean}, \textit{sum}, and \textit{max} aggregators being the most used in the state-of-the-art models \cite{NRI, stochasticGNN, GNN_multi_agent_2}. However, Xu \textit{et al.} \cite{GNN_theory1} showed that \textit{mean} and \textit{max} aggregators are well-defined multiset functions due to their permutation invariant, but they are \textit{not} injective. They proposed to use a \textit{sum} aggregator to distinguish between node neighbourhoods. Furthermore, Corse \textit{et al.} \cite{GNN_theory2} proved that by combining a \textit{mean} aggregator with a linear-degree scaler, the resulting aggregator becomes injective: } 

\noindent\YH{\textbf{Theorem 4} (Injective functions on countable multisets)\textbf{.} \textit{The mean aggregation composed with any scaling linear to an injective function on the neighbourhood size can generate injective functions on bounded multisets of countable elements.}}

\YH{Based on the aforementioned theorems, our proposed SGN model with multiple injective aggregators should have more representational power (i.e. have more capability to infer internal spatial-temporal structural relations) than other state-of-the-art GNN models that either use non-injective aggregator or single injective aggregator. Specifically, the SGN structure contains both the \textit{attention-weighted} aggregator (i.e. Eq.~\ref{eq:aggregator_1}) and the \textit{sum} aggregator (i.e. Eq.~\ref{eq:aggregator_2}). Note that this is a general and flexible architecture, which we applied two neighbour-aggregations. Higher degree graphs such as social networks could benefit from further aggregators. }

\begin{figure}[htbp]
	\centering
	\includegraphics[scale=0.44]{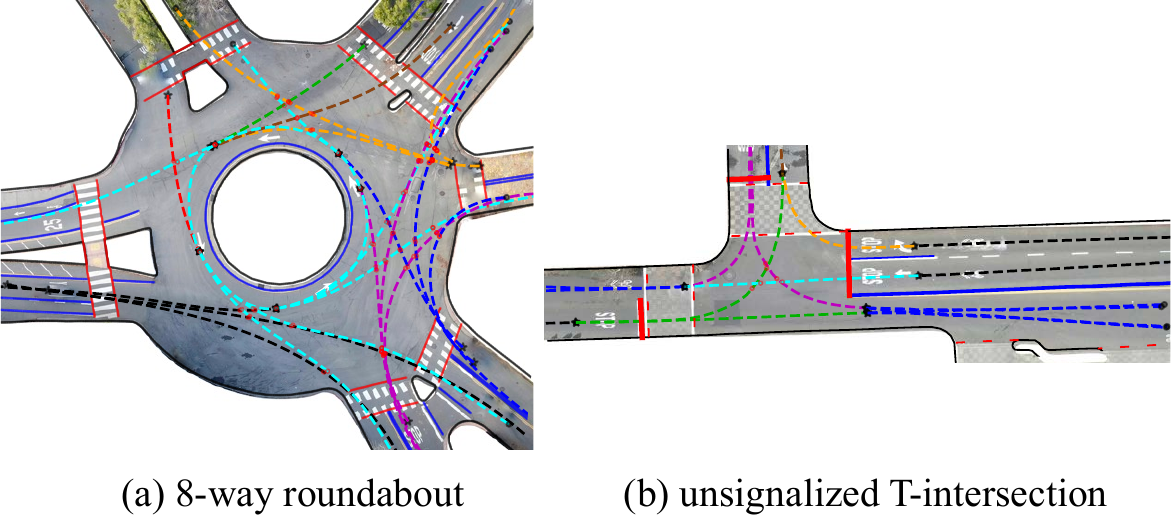}
	\caption{Scenarios that are utilized in this paper as well as their corresponding reference paths.}
	\label{fig:scene_map}
\end{figure}
\section{Experiments on Real-World Scenarios}
\label{experiment}
In this section, we evaluate the capability of the proposed algorithm through different aspects, where its overall performance, flexibility, and transferability are examined. 

\subsection{Dataset}
The experiment is conducted on the INTERACTION dataset \cite{dataset}, \cite{dataset_2}, where two different scenarios are utilized: a 8-way roundabout and an unsignalized T-intersection. All data were collected by a drone from bird's-eye view with 10 Hz sampling frequency. Information such as reference paths and traffic regulations can be directly extracted from high definition maps that are in lanelet \cite{lanelet2} format. We further utilize piece-wise polynomial to fit each of the reference paths in order to improve smoothness. Figure \ref{fig:scene_map} shows the two scenarios we used in this work as well as their reference paths. 

The roundabout scenario is used to evaluate the flexibility and prediction accuracy of the proposed semantic-based algorithm. The intersection scenario, on the other hand, is used to examine the transferability of the algorithm. For roundabout scenario, a total of 21,868 data points are extracted and randomly split into approximately 80\% for training and 20\% for testing. For intersection scenario, there are 13,653 data points in total and we randomly select 80\% of the data to train a new SGN model specifically for intersection scenario. The rest of the data collected at the intersection are used to evaluate the transferability of the SGN model learned under the roundabout scenario.


\begin{figure*}[htbp]
	\centering
	\includegraphics[scale=0.4]{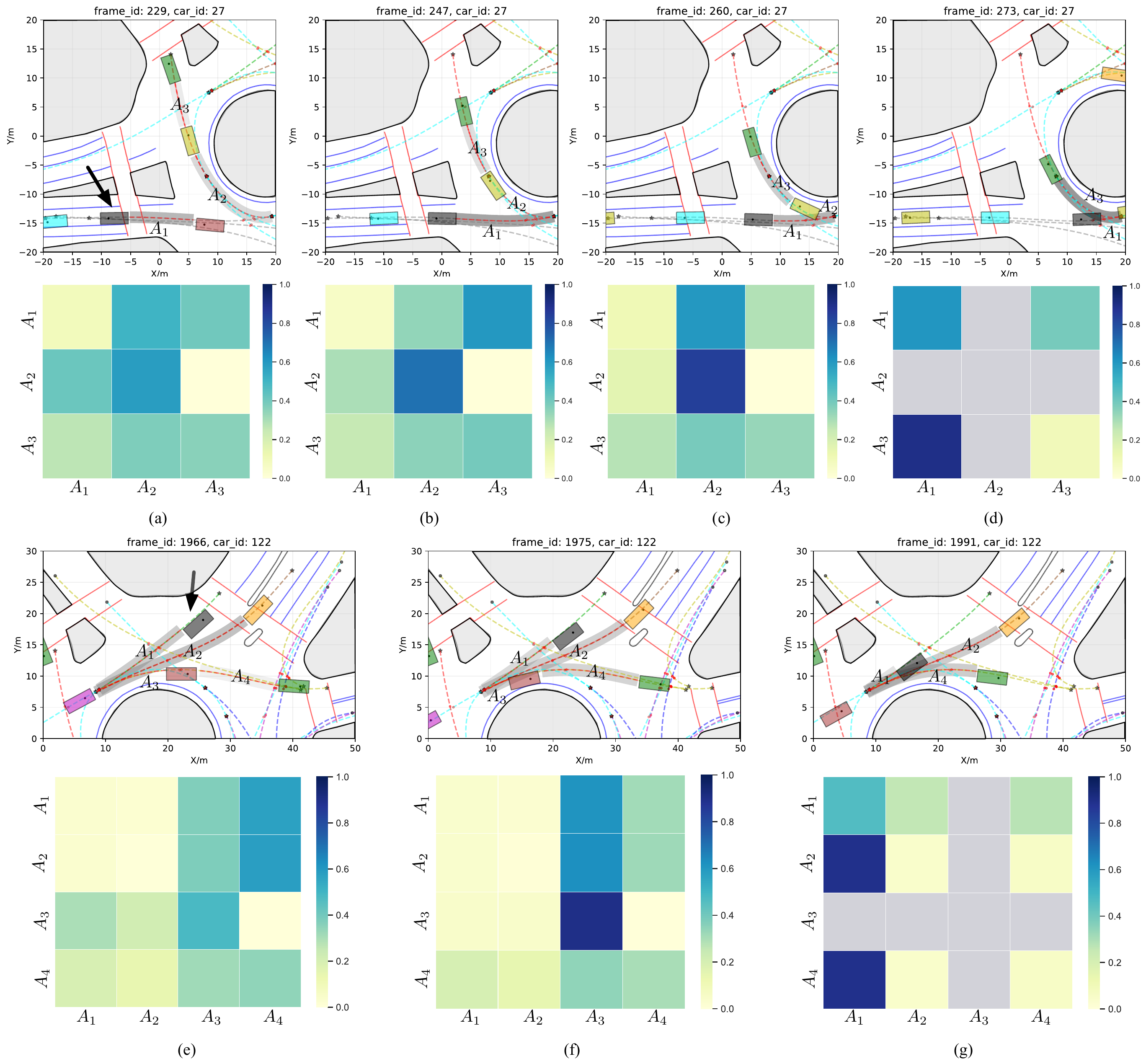}
	\caption{Visualization results of semantic intention and attention heatmap for case 1 (a)-(d) and case 2 (e)-(g). The predicted vehicle is colored in black. The darker the color of the dynamic insertion area, the higher probability for it to be inserted by the predicted vehicle. For each DIA that might be inserted by the predicted vehicle, the corresponding horizontal grids in the heatmap reflect how much its states will be influenced by other DIAs respectively.}
	\label{fig:test_scene_heatmap}
\end{figure*}

\subsection{Implementation Details}
In our experiment, up to two historical time steps of the semantic graph are utilized as input to the network although more historical time steps can be considered to improve the prediction performance. The reason is because, in this work, we focus more on how the proposed generic representations can be integrated into the graph-based network structure to enable flexibility and transferability of the prediction algorithm. Moreover, we want to show that even with limited historical information, the proposed algorithm can still generate desirable results.

The dimension for GRU-based recurrent functions, $f_{rec}^{1,2}$, is set to 128. All the layers in the network embed the input into a 64-dimensional vector with $tanh$ non-linear activation function. A dropout layer is appended to the end of each layer to enhance the network’s generalization ability and prevent overfitting. The size of the attention weight, $\mathbf{W}_{att}$, is set to 128. A batch size of 512 is used at each training iteration with learning rate of 0.001. The model is implemented in PyTorch and it is optimized on 2 GPUs with synchronous training.

\subsection{Visualization Results}
We selected two distinct traffic situations under the roundabout scenario to visualize our test results, where the number of road agents in each case is time varying. The semantic intention prediction results and the corresponding normalized attention coefficients (represented through heatmap) at each tested frame for the two driving cases are shown in Fig.~\ref{fig:test_scene_heatmap}. It should be stressed that the attention coefficients are implicitly learned in the spatial attention layer of SGN during training without any supervision. 

\subsubsection{Case 1}
In Fig.~\ref{fig:test_scene_heatmap}(a)-(d), the predicted vehicle (colored in black) manages to enter the roundabout and, at the meantime, it needs to interact with the other two vehicles that it may have conflict with. At the time frame in Fig.~\ref{fig:test_scene_heatmap}(a), the predicted vehicle begins to enter the roundabout and it has three options: (1) insert into $\mathcal{A}_1$, which can be interpreted as keep following its front car while expecting the other two cars (i.e. the yellow and green vehicle) to pass first; (2) insert into $\mathcal{A}_2$, which is equivalent to cut in front of the yellow vehicle; (3) insert into $\mathcal{A}_3$, which can be regarded as cut in between the green and yellow vehicle. Our result reveals that at such situation, the predicted vehicle has roughly equal probability of inserting into $\mathcal{A}_1$ and $\mathcal{A}_2$, with slightly lower probability of entering $\mathcal{A}_3$. As the predicted vehicle keeps moving forward (Fig.~\ref{fig:test_scene_heatmap}(b) - (d)), its probability of inserting into $\mathcal{A}_2$ decreases and goes to zero while the probability of inserting into $\mathcal{A}_3$ increases.

We also visualize the learned attention coefficients to examine whether the applied attention mechanism learned to associate different weights to different DIAs with reasonable interpretations. According to the attention heatmaps, $\mathcal{A}_1$'s attention vacillates between $\mathcal{A}_2$ and $\mathcal{A}_3$ to decide which area the predicted vehicle will insert into. After the decision is made, $\mathcal{A}_1$ does not need to care about other areas besides itself and thus its own attention coefficient gets higher in Fig.~\ref{fig:test_scene_heatmap}(d). On the other hand, $\mathcal{A}_2$ initially pays some attention to $\mathcal{A}_1$ but it gradually diverse its attention from $\mathcal{A}_1$ after realizing $\mathcal{A}_1$ does not have much interaction with itself. Note that $\mathcal{A}_2$ pays no attention to $\mathcal{A}_3$ throughout the entire period as its future states will not be influenced by its rearward DIAs. The insertion area $\mathcal{A}_3$ uniformly assign its attention to all DIAs until it is about to be inserted by the predicted vehicle where $\mathcal{A}_3$ starts to pay more attention to $\mathcal{A}_1$.



\subsubsection{Case 2}
\begin{figure}[htbp]
    \centering
	\includegraphics[scale=1.1]{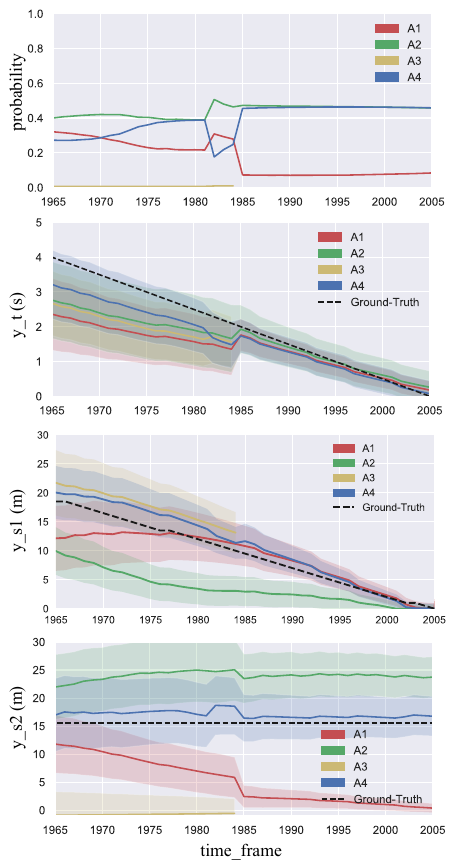}
	\caption{Illustration of the behavior prediction results for test case 2.  For each DIAs that might be inserted by the predicted vehicle and for each goal-state variable, we plotted the predicted mean value and confidence interval based on fifty samples.}
	\label{fig:test_result_122}
\end{figure}

Different from case 1, this driving case is a situation where the predicted vehicle has to interact with vehicles driving on different reference paths while entering the roundabout (Fig.~\ref{fig:test_scene_heatmap}(e)-(g)). Initially, the predicted vehicle can choose to insert into either one of the four areas (i.e. $\mathcal{A}_1$, $\mathcal{A}_2$, $\mathcal{A}_3$, $\mathcal{A}_4$). In Fig.~\ref{fig:test_scene_heatmap}(e), inserting into $\mathcal{A}_3$ has zero probability for the predicted vehicle since the area size is small and it is hard to be reached. Inserting into $\mathcal{A}_4$ also has low probability since the predicted vehicle has large geometric distance to $\mathcal{A}_4$ at the current time step. As the predicted vehicle keeps moving forward (Fig.~\ref{fig:test_scene_heatmap}(f), (g)), the probabilities of inserting into $\mathcal{A}_2$ and $\mathcal{A}_4$ increase and almost equal to each other. Eventually, the predicted vehicle inserted into both $\mathcal{A}_2$ and $\mathcal{A}_4$. The corresponding attention heatmaps also provide some reasonable interpretations of this driving case. For example, $\mathcal{A}_1$ and $\mathcal{A}_2$ gradually shift their attention from $\mathcal{A}_4$ to $\mathcal{A}_3$ as they are being inserted by the red car (which formulates the rear bound of $\mathcal{A}_3$). Also, $\mathcal{A}_3$ stops concentrating on $\mathcal{A}_1$ and $\mathcal{A}_2$ as soon as it reveals higher chances to pass the reference point first.

\begin{table*}[!ht]
	\begin{center}
		\caption{Quantitative Evaluation Results. As mentioned in Section V.D, $y_t$ denotes the time remains for the predicted vehicle to finish the insertion; $y_{s_1}$ denotes the location of the inserted DIA; and $y_{s_2}$ denotes the location of the predicted vehicle relative to the DIA it enters. }
		\label{tab:quantitative_result}
		\resizebox{\textwidth}{!}{
		\begin{tabular}{c||c|c|c|c|c|c|c||c|c|c}
			\toprule
			\multicolumn{1}{c||}{} &
			\multicolumn{7}{c||}{Baseline Methods} &
			\multicolumn{3}{c}{\textbf{SGN (Ours)}} \\
			
			\midrule
			\backslashbox{ \quad }{ \quad \quad  } & \thead{MC-dropout \\ \cite{dropout}} & \thead{SIMP \\\cite{SIMP}}  & \thead{DESIRE \\ \cite{CVAE_DESIRE}} & \thead{SGAN \\ \cite{social_GAN}} & \thead{NRI \\ \cite{NRI}} & \thead{SGNN \\ \cite{stochasticGNN}} & \thead{TrafficPredict \\ \cite{GNN_multi_agent_2}} & NC & UA & \thead{CA}\\
			\midrule
			Prob ($\%$) & 83.52 & 91.29 & 90.04 & 85.12 & 92.77 & 93.01 & 93.59 & 93.62  & 95.05 & \textbf{95.87} \\
			\midrule
			 Time - $y_t$ (s) &  2.11 $\pm$ 0.02  & 1.07 $\pm$ 1.06 & 1.75 $\pm$ 0.05 & 1.97 $\pm$ 0.34 & 1.10 $\pm$ 0.93 & 1.07 $\pm$ 0.89 & 1.03 $\pm$ 0.79 & 1.02 $\pm$ 0.78 & 1.02 $\pm$ 0.84 & \textbf{0.95 $\pm$ 0.79} \\
			\midrule
			 Loc 1 - $y_{s_1}$ (m) &  5.80 $\pm$ 0.70 & 4.51 $\pm$ 4.66 &  6.93 $\pm$ 4.19 & 6.81 $\pm$ 4.34 & 4.68 $\pm$ 4.21 & 3.67 $\pm$ 4.15 & 3.51 $\pm$ 4.10 & 5.88 $\pm$ 6.05 & 3.87 $\pm$ 4.26 & \textbf{3.45 $\pm$ 4.06}\\
			\midrule
			 Loc 2 - $y_{s_2}$ (m) & 6.35 $\pm$ 1.99 & 5.02 $\pm$ 5.16 &  5.75 $\pm$ 4.65 & 5.72 $\pm$ 4.70 & 5.01 $\pm$ 4.54 & 3.71 $\pm$ 4.44 & 3.63 $\pm$ 4.38 & 4.69 $\pm$ 5.05 & 3.84 $\pm$ 4.53 & \textbf{3.55 $\pm$ 4.25}\\		
			\bottomrule
		\end{tabular}}
	\end{center}
\end{table*}

We further illustrate numerical results of semantic intention and semantic goal state prediction for all possible insertion areas at each time step of this driving case, which are shown in Fig.~\ref{fig:test_result_122}. It is worth to note that both $\mathcal{A}_2$ and $\mathcal{A}_4$ can be regarded as the final insertion area for this case, but $\mathcal{A}_4$ is chosen since its rear bound is closer to the predicted vehicle than that of $\mathcal{A}_2$. The first plot shows the insertion probability of each DIA at each time step, the value of which coincide with the visualization results in Fig.~\ref{fig:test_scene_heatmap}(e)-(g). As can be seen from the last three plots in Fig.~\ref{fig:test_result_122}, each predicted state for $\mathcal{A}_4$ does not have large deviation from the ground truth in terms of the mean value. Also, the variance of each predicted state gradually decreases as the predicted vehicle gets closer to finish insertion. Moreover, even for those dynamic insertion areas that are not eventually inserted by the predicted vehicle, our proposed algorithm is still able to make reasonable predictions. 


\subsection{Qualitative Result Evaluation}
We compared the performance of our model with that of the following seven state-of-the-art methods approaches: MC-dropout\cite{dropout}, SIMP\cite{SIMP}, DESIRE\cite{CVAE_DESIRE}, SGAN\cite{social_GAN}, NRI\cite{NRI}, SGNN\cite{stochasticGNN}, TrafficPredict\cite{GNN_multi_agent_2}. For a fair comparison, we keep the output and input representations the same, which are based on the proposed semantic-based representations. Moreover, hyper-parameters such as the number of neurons, batch size, dropout rate, and training iterations in all these methods are also kept the same. 

We also conduct several ablation studies to investigate contribution of two components to our method’s performance: concatenation (\textbf{C}) and spatial attention (\textbf{A}). With no concatenation (\textbf{NC}), the model is the proposed method with modification on the predictor encoding layer. Specifically, for Eq.(9), we directly use $\hat{\mathbf{h}}_i^t$ as input for $f_{enc}^2$, instead of concatenating it with $\hat{\mathbf{h}}_{j \rightarrow i}^t$. In this way, the input of $f_{pred}$ becomes the hidden edge relation between node $i$ and $j$. With uniform attention (\textbf{UA}), it is equivalent to the proposed method with modification on the spatial attention layer, where we manually assign uniform attention coefficient to each node. When both concatenation and spatial attention are considered (\textbf{CA}), the model becomes the proposed SGN.

The intention prediction results are evaluated by calculating the multi-class classification accuracy and the semantic goal state prediction results are evaluated using root-mean-squared-error (RMSE) as well as standard deviation. Note that the input dimension has to be fixed for baseline models due to the limitation of their network structures. Therefore, only a fixed number of surrounding DIAs can be considered. As most scenarios have three surrounding DIAs, we select three DIAs that are closest to the predicted vehicle to extract input features for baseline methods. If less than three surrounding DIAs exist at a certain time frame, we assign features of those non-existent DIAs to zero. 

According to the results shown in Table~\ref{tab:quantitative_result}, MC-dropout has the lowest intention prediction accuracy and the smallest prediction variance amongst all baseline methods. This is because the dropout method is incapable of bringing enough uncertainties to the model and thus it is more likely to have over-fitting problems. Moreover, DESIRE and SGAN have slightly worse performances than SIMP, which might due to the loss terms of the two methods. In fact, the loss function for DESIRE has a trade-off between a good estimation of data and the Kullback–Leibler (KL) divergence for latent space distribution, which two terms need to be carefully fine-tuned for desirable results. Similarity, for SGAN, the generator loss can lead to the GAN instability, which can cause the gradient vanishing problem when the discriminator can easily distinguish between real and fake samples. For the three graph-based approaches, TrafficPredict has the best performance compare to NRI and SGNN. Although these graph-based methods exhibit better prediction accuracy compared to the other baseline methods, their performances are still worse than the proposed method. \YH{This is because, as mentioned in Section \ref{SGN_theory}, these graph-based methods are only suitable for problems where each node is defined by a single agent, which do not take into consideration of intra-node relations. Therefore, when each node is defined by semantic representation that contains both context and agent information, typical graph networks fail to accurately reason internal relations among these semantic graphs as they either use non-injective aggregator or single injective aggregator. Specifically, NRI only uses \textit{sum} aggregator, while SGNN and TrafficPredict only use \textit{mean} aggregator.}

Among all the ablation methods, NC has the worst overall performance, which shows the necessity of emphasizing the relations between features of the reference DIA and other DIAs. The test results of our proposed method surpass those of UA in terms of the prediction accuracy, which implies the advantages of using attention mechanism to treat surrounding DIAs with different importance. Also, as the prediction results of all three SGN-based methods outperforms those of the three baseline methods, we can conclude that utilizing graph-based networks are, in general, better than traditional learning-based methods that have weak inductive bias. Specifically, the flexibility of dealing with varying number of input features as well as the invariance to feature ordering are essential properties for relational reasoning under prediction tasks.  

\begin{figure*}[htbp]
	\centering
	\includegraphics[scale=0.77]{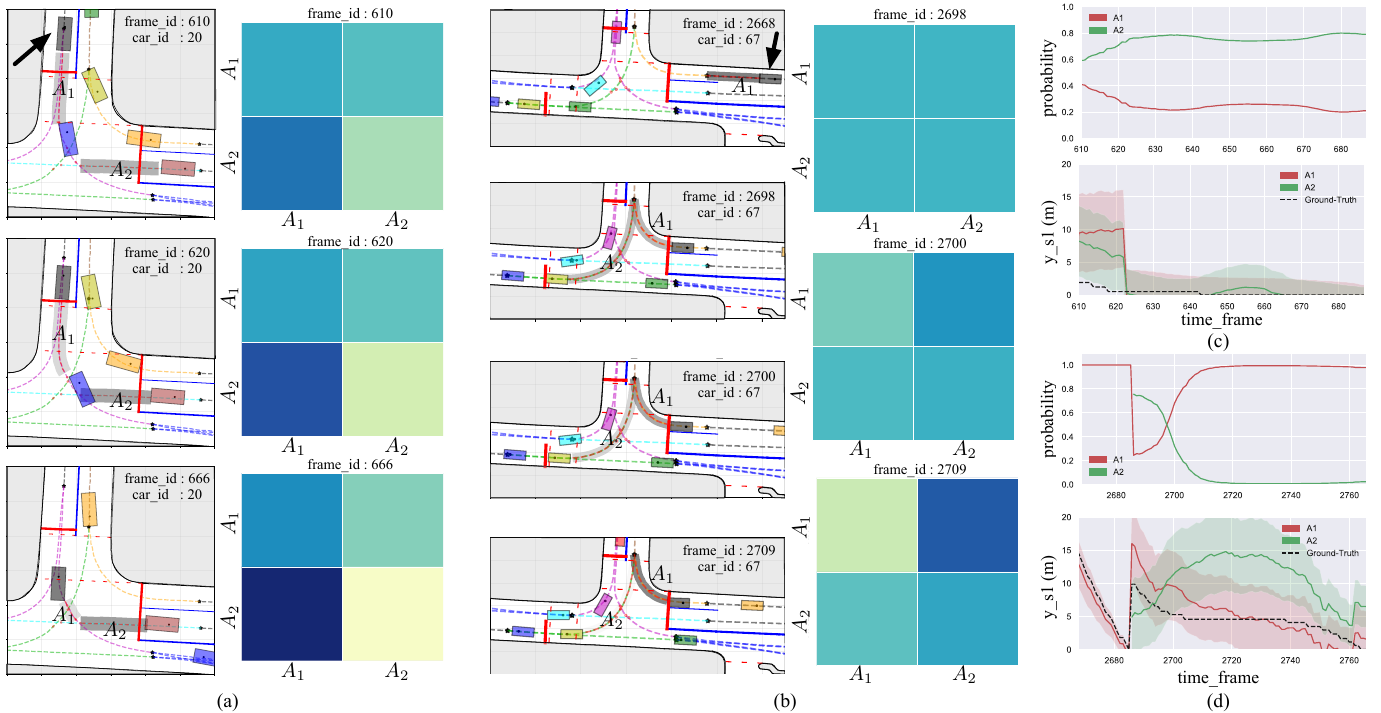}
	\caption{Visualization results of semantic intention and attention heatmap for test case 3 (a) and test case 4 (b). Also, selected behavior prediction results for case 3 (c) and case 4 (d) are also illustrated. All these results are generated by the zero-shot transferred SGN model.}
	\label{fig:adapt_test_result_all}
\end{figure*}

\subsection{Analysis of Scene Transferability}
\label{scene_transfer}
As mentioned in Section II.D, our proposed prediction algorithm is intended to have zero-shot transferability. In this subsection, we explicitly examine how our model, trained under a single domain, is performed when tested under a completely unforeseen domain. To be more specific, we trained our model under an 8-way roundabout and test it under an unsignalized T-intersection. 


\subsubsection{Overall qualitative evaluation}
As mentioned in Section VI.A, to evaluate the transferability of the proposed algorithm, the prediction performances of two SGN models are compared using selected test data from the intersection: (1) the first SGN model is learned using only the training data from the roundabout scenario, which is the same model we used for previous evaluations; (2) the second SGN model is learned using only the training data from the intersection scenario. It should be emphasized that the first SGN model is directly tested without additional training on the intersection data. Therefore, we name the first model as the \textit{zero-shot transferred model}. In contrast, the second SGN model is named as the \textit{conventional model}. The testing results are shown in Table~\ref{tab:transfer}.

\begin{table}[!ht]
	\begin{center}
		\caption{Evaluation of Transferability}
		\label{tab:transfer}
	\begin{tabular}{c|c|c}
		\toprule
		\multicolumn{1}{c|}{} &
		\multicolumn{1}{c|}{Zero-shot Transferred Model} &
		\multicolumn{1}{c}{Conventional Model} \\
		\midrule
		Prob (\%) & 93.73 & 94.68\\
		\midrule
		Time - $y_t$ (s) & 2.12 $\pm$ 0.79 & 1.49 $\pm$ 1.55\\
		\midrule
		Loc 1 - $y_{s_1}$ (m) & 4.24 $\pm$ 3.86 & 2.67 $\pm$ 2.13\\
		\midrule
		Loc 2 - $y_{s_2}$ (m) & 4.87 $\pm$ 4.13 & 1.41 $\pm$ 2.71\\
		\bottomrule
	\end{tabular}
	
	\end{center}
\end{table}

From the table, we first notice that the performances of the conventional model using the proposed SGN structure are satisfying in terms of the prediction accuracy. More precisely, the intention prediction accuracy is close to 95\% , the average temporal estimation of the goal state is less than 1.5s, the average estimation of the semantic goal location is around 2m, and the variances of these predicted variables are within a reasonable range. When the results of these two models are compared, we observe that the performance of the zero-shot transferred model still maintain desirable performance compared to the conventional model. Specifically, the semantic intention prediction accuracy only decreases 1\% , the average temporal prediction error increases 0.5s, and the mean estimation error for semantic goal locations rises 2m.

\subsubsection{Case studies}
Two testing cases in the intersection scenario are selected to provide visualization results and detailed analysis. It is worth emphasizing that the testing results shown below are all generated through the zero-shot transferred SGN model learned under the roundabout scenario. The differences between these two domains are mainly related to map information (e.g. road topology) and traffic situation (e.g. number of on-road vehicles). 

\noindent\textbf{Case 3}: Figure ~\ref{fig:adapt_test_result_all}(a) is a case where two vehicles reach the stop line simultaneously and they need to negotiate the road with each other. According to the results in Fig.~\ref{fig:adapt_test_result_all}(a) and (c), the transferred model is able to successfully infer the semantic intention of the predicted vehicle at an early stage (i.e. 7s before it finally inserts into $\mathcal{A}_2$). According to the corresponding heatmaps, the state of $\mathcal{A}_2$ have less effects on the predicted vehicle's decision than the state of $\mathcal{A}_1$. This is because there is no much change on the state of $\mathcal{A}_2$ and thus the predicted vehicle infers that it is unnecessary to pay much attention to $\mathcal{A}_2$. The second plot in Fig.~\ref{fig:adapt_test_result_all}(c) is the predicted result of $y_{s_1}$ for each DIA. Since the red vehicle keeps waiting behind the stop line, the ground truth of $s_1$ for $\mathcal{A}_2$ is close to zero during this period. According to the plot, our transferred model successfully predicts such behavior with relatively small variance.
    
\noindent\textbf{Case 4}: Figure ~\ref{fig:adapt_test_result_all}(b) is a driving case that consists of two different stages, where the predicted vehicle first need to drive towards the stop line and then make a right turn. When the predicted vehicle is approaching the stop line, the only available insertion area is $\mathcal{A}_1$. Hence, during the first stage, the probability of inserting into $\mathcal{A}_1$ remains at one and our transferred predictor successfully infers the state changes of $\mathcal{A}_1$ as shown in Fig.~\ref{fig:adapt_test_result_all}(d). When the predicted vehicle is close to the stop line and preparing for a right turn, it notices that a yellow car is turning left and has potential conflict with itself. According to the first plot in Fig.~\ref{fig:adapt_test_result_all}(d), the inserting probability of $\mathcal{A}_1$ gradually increases (i.e. the possibility for the predicted vehicle to yield increases) and about 6s before the final insertion, the predictor is certain that $\mathcal{A}_1$ is the ground-truth DIA. From the second plot of Fig.~\ref{fig:adapt_test_result_all}(d), although the ground truth of $s_1$ changes non-linearly with time, our transferred model are still able to make relatively precise predictions. Opposite from case 3, the state of $\mathcal{A}_1$ has less variances than that of $\mathcal{A}_2$ and thus the predicted vehicle's decision depends more on $\mathcal{A}_2$. The intuition is the predicted vehicle needs to keep track of $\mathcal{A}_2$'s state in order to decide when the right turn can be made.

\subsubsection{Discussion and further impact}  In fact, since both driving scenarios we considered are in urban area with the same speed limit, without the loss of generality, we assume that the overall driving styles under these two domains are similar (i.e. have similar distributions). Therefore, in our case, the train-test domain shift 
\begin{figure}[htbp]
    \centering
	\includegraphics[scale=0.4]{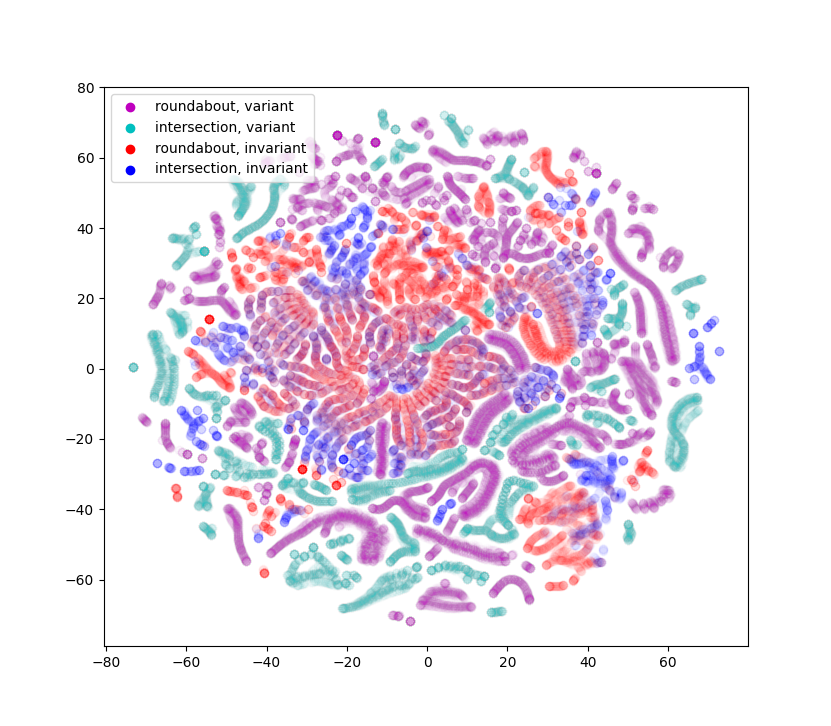}
	\caption{The t-SNE plots for visualizing representations of proposed \textit{generic representation} (colored in \textcolor{blue}{blue} and \textcolor{red}{red}) and \textit{typical representation} (colored in \textcolor{cyan}{cyan} and \textcolor{magenta}{magenta}) under different domains.}
	\label{fig:other_analysis}
\end{figure}
is mainly due to distinct map information
and different inter-vehicle relations. However, if the driving style of the test domain is different from that of the training domain (e.g. two domains belong to different countries), zero-shot transfer may not have desirable performance and we may need an extra step to align the two domain distributions, which will be considered in our future works.

In general, after the proposed model is offline trained using data collected from limited driving scenarios, it can be directly utilized online under unforeseen driving environments that have different road structures, traffic regulations, number of on-road agents, and agents' internal relations. Moreover, the proposed method is data-efficient since when a new scenario is encountered, no extra data have to be collected to re-train the model for prediction tasks. Indeed, when an autonomous vehicle is navigating in constantly changing environments, it might be incapable of collecting enough online data to train a new predictor under each upcoming scenario. 
\YH{\subsection{Representation Analysis}}
\label{additional_analysis}
We utilize t-SNE \cite{tsne} to analyze whether the proposed semantic graph has shown to have similar marginal distribution across domains. Specifically, for every roundabout and intersection data, we embed graph nodes features in \textit{generic representation} and in \textit{typical representation} (i.e. using Cartesian coordinate) into a two-dimensional space, respectively. It appears from Fig.~\ref{fig:other_analysis} that our generic representations under two different domains (i.e. intersection and roundabout) achieve a higher overlap compared to traditional representation. The result shows that by processing static and dynamic environment information through proposed pipelines, the resulted representations are invariant across domains. Consequently, prediction models using such representations possesses zero-shot transferability, which is proved both theoretically in Section~\ref{SG_theory} and empirically in Section~\ref{scene_transfer}.

\section{Conclusion}
In this paper, a scenario-transferable semantic graph reasoning approach is proposed for interaction-aware vehicle behavior prediction. A generic environment representation that takes the advantage of semantics and domain knowledge is proposed. These representations are utilized to define semantic goals which are then integrated with the concept of semantic graphs to construct structural relations within these representations. Finally, by proposing the semantic behavior prediction framework to operate on semantic graphs, the overall prediction framework not only achieves state-of-the-art performances, but also exhibits great transferability to unforeseen driving scenarios with completely different road structures and traffic regulations. Specifically, in our experiments, we first utilize two representative scenarios to visually illustrate the prediction performance of the algorithm. We then thoroughly evaluate the algorithm under a real-world scenario. According to the results, our method outperforms seven baseline methods in terms of both the prediction error and the confidence intervals. We also evaluate the predictor's performance of directly transferring the predictor learned in an 8-way roundabout to an unsignalized T-intersection. The result shows that by directly extracting interpretable domain-invariant representations based on prior knowledge and incorporating these representations into the semantic graph structure as input, the proposed prediction architecture achieves desired transferability or domain generalizability. Moreover, by using graph networks that operate on these graphs and reason internal pairwise structural relations, the proposed algorithm is also invariant to the number and order of input features, which further enables transferability of the predictor. 

For future works, we will consider heterogeneous agents (e.g. pedestrians and cyclists) in the environment and other types of domain shift to improve transferability of the predictor. We will also use the predicted semantic goal state information for downstream tasks such as trajectory prediction and goal-based planning. 

\ifCLASSOPTIONcaptionsoff
  \newpage
\fi

\bibliographystyle{IEEEtran.bst}
\bibliography{IEEEabrv, reference}

\begin{IEEEbiography}[{\includegraphics[width=1.0in,height=1.25in, clip,keepaspectratio]{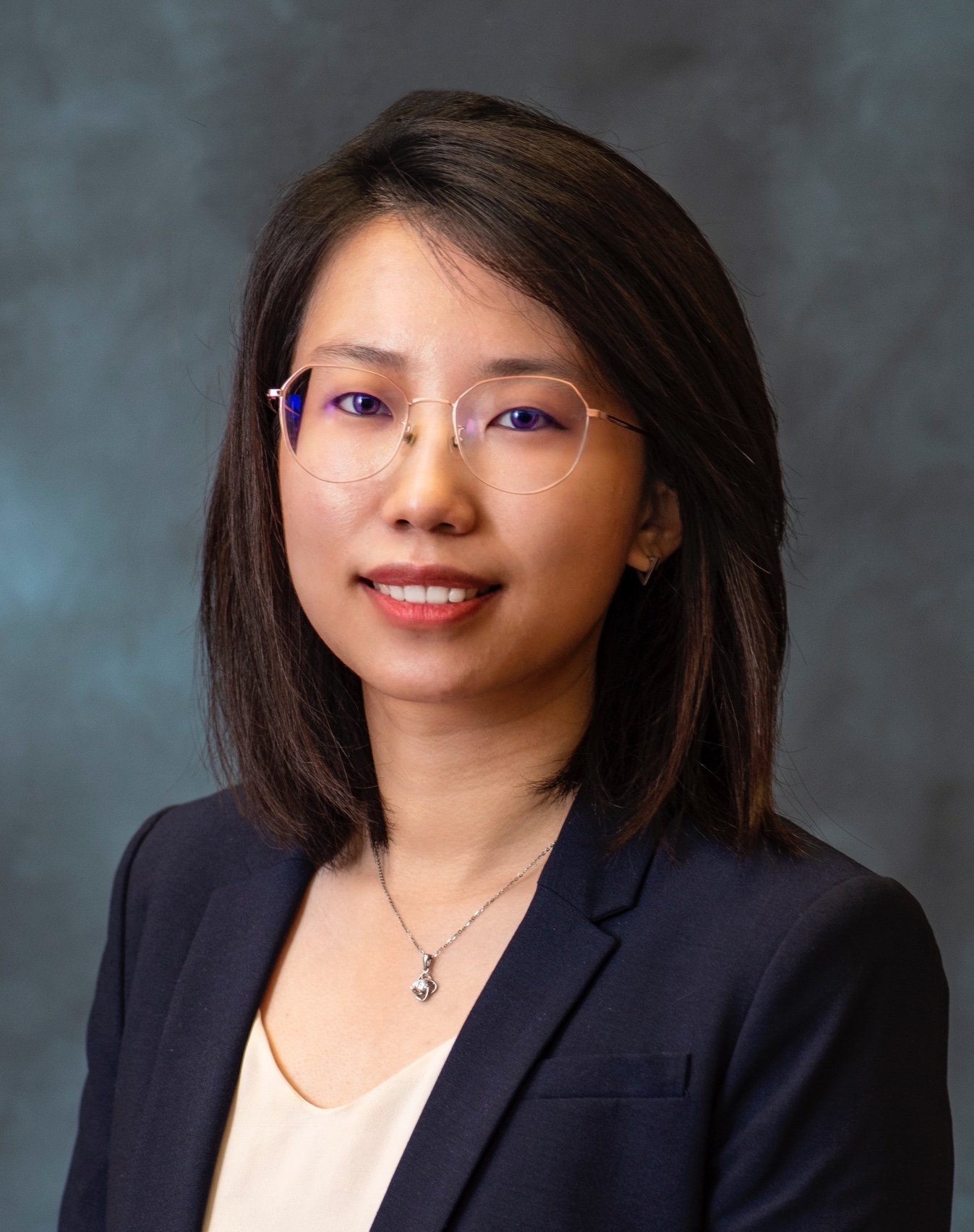}}]{Yeping Hu}
received her Ph.D. degree in Mechanical Engineering from  University of California, Berkeley in 2021. She is currently a postdoctoral researcher at Lawrence Livermore National Laboratory. Her research interests include machine learning, probabilistic models, optimization, reinforcement learning and their applications to behavior prediction, decision making, and motion planning for mobile robots such as intelligent vehicles. She served as an Associate Editor in IEEE IV from 2019 to 2022. She is the recipient of Best Paper Award Finalist in IEEE/RSJ IROS 2019 and Best Student Paper Award in IEEE IV 2018. 
\end{IEEEbiography}

\begin{IEEEbiography}[{\includegraphics[width=1.0in,height=1.298in, clip]{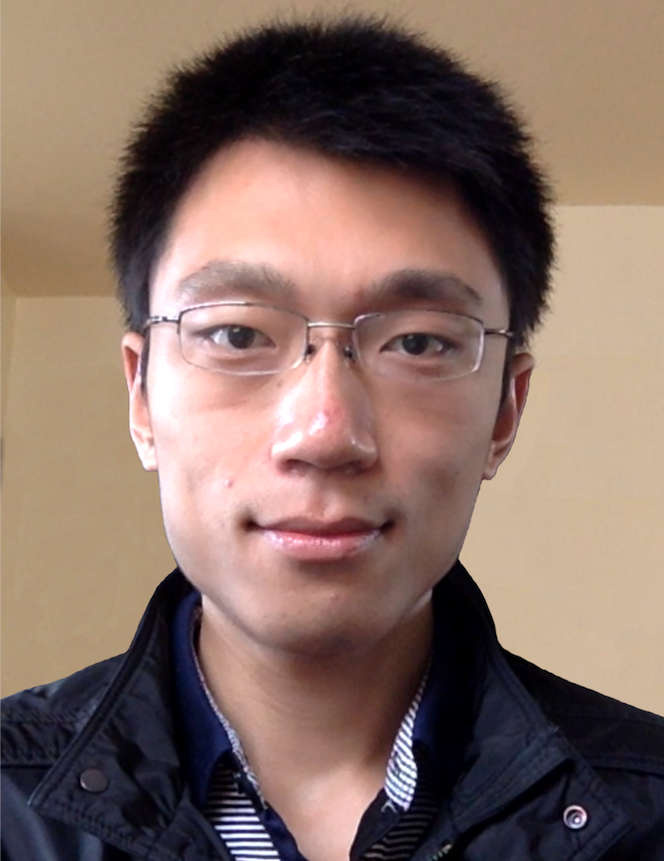}}]{Wei Zhan}
received his Ph.D. degree from University of California, Berkeley in 2019. He is currently an Assistant Professional Researcher at UC Berkeley, and Co-Director of Berkeley DeepDrive Center. His research has been targeting scalable and interactive autonomy at the intersection of computer vision, machine learning, robotics, control and intelligent transportation. His publications received Best Student Paper Award in IV 2018 and Best Paper Award – Honorable Mention in IEEE Robotics and Automation Letters. He is the lead author of the INTERACTION dataset, and organized its prediction challenges in NeurIPS 2020 and ICCV 2021.
\end{IEEEbiography}

\begin{IEEEbiography}[{\includegraphics[width=1.04in,height=1.23in,clip]{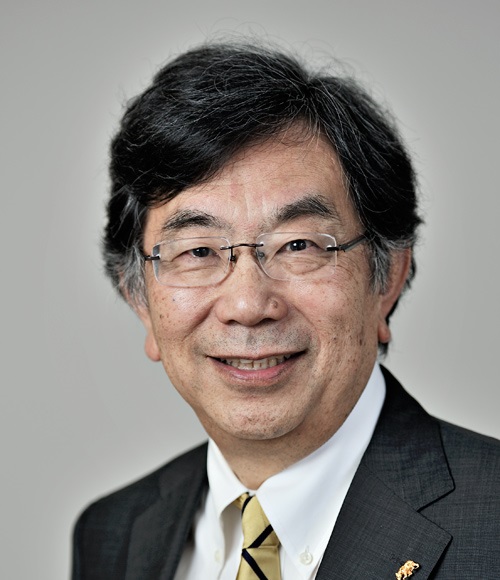}}]{Masayoshi Tomizuka}
 (M’86-SM’95-F’97-LF’17) received his Ph. D. degree in Mechanical Engineering from MIT in February 1974. In 1974, he joined the faculty of the Department of Mechanical Engineering at the University of California at Berkeley, where he currently holds the Cheryl and John Neerhout, Jr., Distinguished Professorship Chair. His current research interests are optimal and adaptive control, digital control, signal processing, motion control, and control problems related to robotics, precision motion control and vehicles. He served as Program Director of the Dynamic Systems and Control Program of the Civil and Mechanical Systems Division of NSF (2002- 2004). He served as Technical Editor of the ASME Journal of Dynamic Systems, Measurement and Control, J-DSMC (1988-93), and Editor-in-Chief of the IEEE/ASME Transactions on Mechatronics (1997-99). Prof. Tomizuka is a Fellow of the ASME, IEEE and IFAC. He is the recipient of the Charles Russ Richards Memorial Award (ASME, 1997), the Rufus Oldenburger Medal (ASME, 2002) and the John R. Ragazzini Award (2006).
\end{IEEEbiography}

\end{document}